\documentclass[10pt,journal,compsoc]{IEEEtran}

\ifCLASSOPTIONcompsoc
  \usepackage[nocompress]{cite}
\else
  \usepackage{cite}
\fi

\usepackage{amsmath}
\usepackage{amssymb}
\usepackage{array}
\usepackage{bbm}
\usepackage{booktabs}
\usepackage{caption}
\usepackage{graphicx}
\usepackage{microtype}
\usepackage{hyperref}
\usepackage{url}
\usepackage{cleveref}
\usepackage{algorithm}
\usepackage{algpseudocode}
\usepackage{color}
\usepackage{multirow}
\usepackage{soul}
\makeatletter
\renewcommand{\ALG@name}{Method summary}
\makeatother
\newcommand{\rev}[1]{\textcolor{black}{{#1}}}

\begin{document} 

\title{AutoNovel: Automatically Discovering and Learning Novel Visual Categories}

\author{Kai Han,
        Sylvestre-Alvise Rebuffi,
        S\'ebastien Ehrhardt,
        Andrea Vedaldi, 
        Andrew Zisserman%
\IEEEcompsocitemizethanks{
\IEEEcompsocthanksitem The authors are with Visual Geometry Group, Department of Engineering Science, University of Oxford, Oxford, OX1 3PJ, United Kingdom. \protect\\
E-mail: \{khan, srebuffi, hyenal, vedaldi, az\}@robots.ox.ac.uk\hfil\break
Corresponding author: Kai Han.
}%
\thanks{}}

\markboth{IEEE TRANSACTIONS ON PATTERN ANALYSIS AND MACHINE INTELLIGENCE}%
{Han \MakeLowercase{\textit{et al.}}: AutoNovel: Automatically Discovering and Learning Novel Visual Categories}

\IEEEtitleabstractindextext{%
\begin{abstract}
We tackle the problem of discovering novel classes in an image collection given labelled examples of other classes.
We present a new approach called \emph{AutoNovel}  to address this problem by combining three ideas:
(1) we suggest that the common approach of bootstrapping an image representation using the labelled data only introduces an unwanted bias, and that this can be avoided by using self-supervised learning to train the representation from scratch on the union of labelled and unlabelled data;
(2) we use ranking statistics to transfer the model's knowledge of the labelled classes to the problem of clustering the unlabelled images;
and, (3) we train the data representation by optimizing a joint objective function on the labelled and unlabelled subsets of the data, improving both the supervised classification of the labelled data, and the clustering of the unlabelled data.
\rev{Moreover}, we propose a method to estimate the number of classes for the case where the number of new categories is not known a priori. We evaluate \emph{AutoNovel}  on standard classification benchmarks and substantially outperform current methods for novel category discovery.
In addition, we also show that \emph{AutoNovel}  can be used for fully unsupervised image clustering, achieving promising results.
\end{abstract}

}

\maketitle

\IEEEdisplaynontitleabstractindextext

\IEEEpeerreviewmaketitle

\IEEEraisesectionheading{\section{Introduction}\label{sec:intro}}
\IEEEPARstart{M}{odern}  machine learning systems can match or surpass human-level performance in tasks such as image classification~\cite{deng09imagnet}, but at the cost of collecting large quantities of annotated training data.
Semi-supervised learning (SSL)~\cite{oliver2018realistic} can alleviate this issue by mixing labelled with unlabelled data, which is usually much cheaper to obtain.
However, these methods still require some annotations for each of the classes that one wishes to learn.
We argue that this is not always possible in real applications.
For instance, consider the task of recognizing products in supermarkets;
thousands of new products are introduced in stores every week, and it would be very expensive to annotate them all.
However, new products do not differ drastically from the existing ones, so prior knowledge of older products should help to discover new products automatically as they arise in the data.
Unfortunately, machines are still unable to effectively detect and learn new classes without manual annotations.

\begin{figure}
\centering
\includegraphics[width=0.75\linewidth]{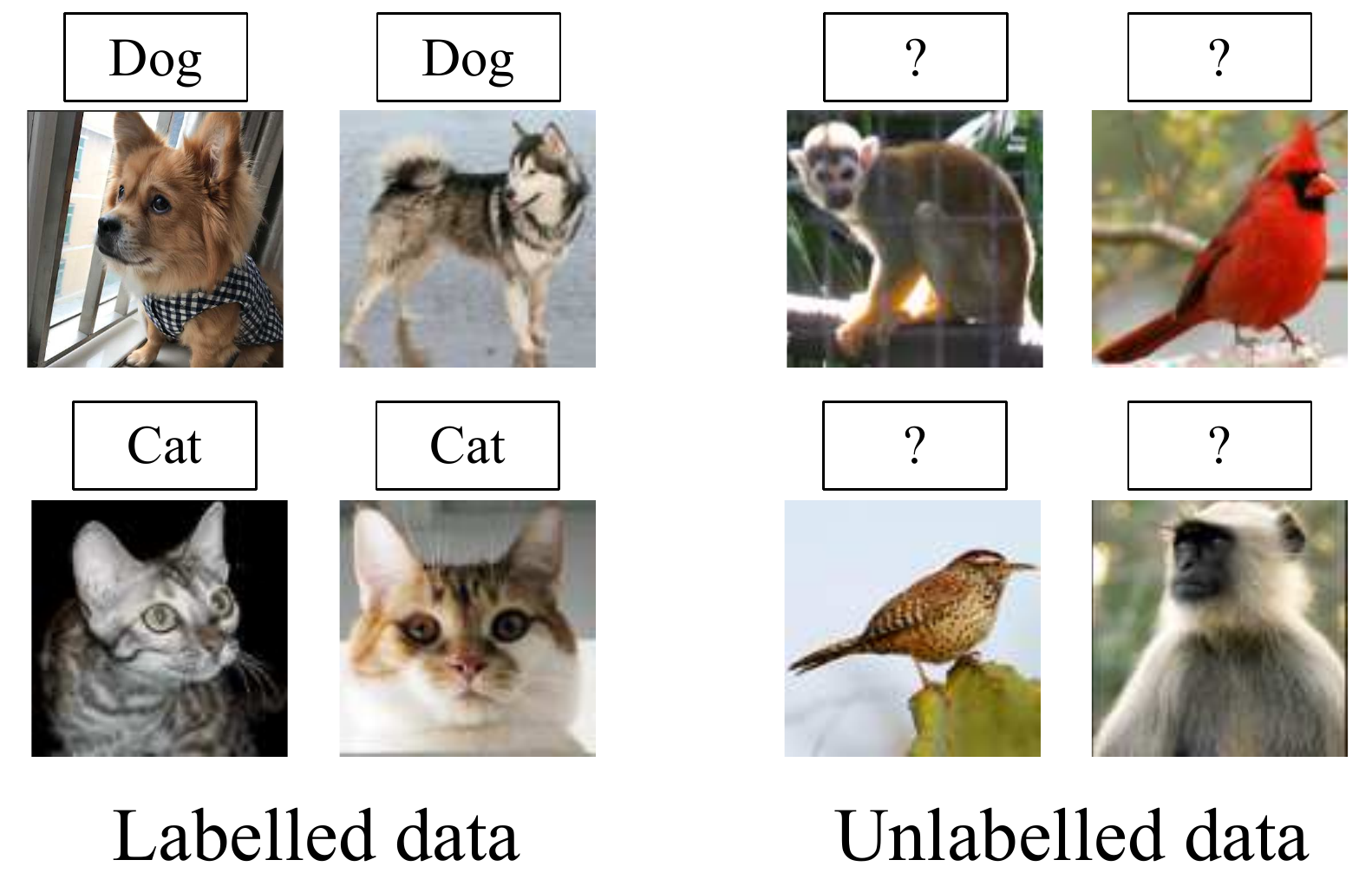}
\caption{Novel category discovery. Given labelled images from a few known categories (e.g., dog and cat), our objective is to automatically partition unlabelled images from new categories (e.g., monkey and bird) into proper clusters.}\label{fig:labelled_unlabelled}
\end{figure}

In this paper, we thus consider the problem of discovering new visual classes automatically, assuming that a certain number of classes are already known by the model~\cite{Hsu18_L2C,Hsu19_MCL,han2019learning} (see~\cref{fig:labelled_unlabelled}).
This knowledge comes in the form of a \emph{labelled dataset} of images for a certain set of classes.
Given that this data is labelled, off-the-shelf supervised learning techniques can be used to train a very effective classifier for the known classes, particularly if Convolutional Neural Networks (CNNs) are employed.
However, this does not mean that the learned features are useful as a representation of the \emph{new classes}.
Furthermore, even if the representation transfers well, one still has the problem of identifying the new classes in an unlabelled dataset, which is a clustering problem.

\begin{figure*}
\centering
\includegraphics[width=0.9\linewidth]{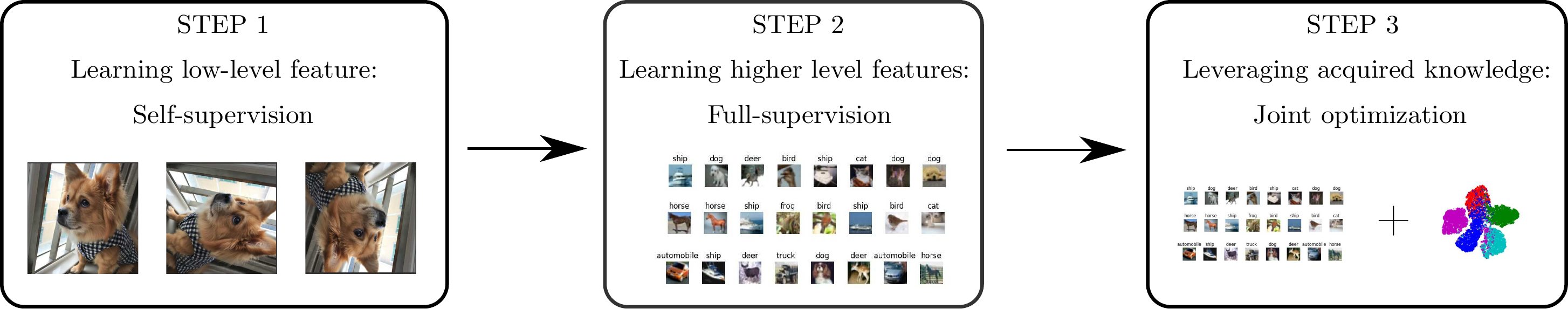}
\caption{Overview of the AutoNovel learning pipeline for novel category discovery.
The first step is to learn an unbiased image representation via self-supervision using both labelled and unlabelled data.
This learns well the early layers of the representation.
The second step is to fine-tune only the last few layers of the representation using supervision on the labelled subset of the data.
The final step is to use the fine-tuned representation, via ranking statistics, to induce clusters in the unlabelled data, while maintaining a good representation on the labelled set.}\label{fig:splash}
\end{figure*}

We tackle these problems by introducing a novel approach called AutoNovel that combines three key ideas (\cref{sec:method,fig:splash}).
The first idea is to pre-train the image representation (a CNN) using all available images, both labelled and unlabelled, using a self-supervised learning objective.
Crucially, this objective \emph{does not} leverage the known labels, resulting in features that are much less biased towards the labelled classes. \rev{Labels are used only after pre-training to learn a classifier specific to the labelled data as well as to fine-tune the last layers of the CNN}.

The second idea is a new approach to transfer the information contained in the labelled images to the problem of clustering the unlabelled ones.
\rev{Information is transferred by sharing the same representation between labelled and unlabelled images, in order to be able to reuse discriminative features learned on the labelled set.}
In more detail, pairs of unlabelled images are compared via their representation vectors.
The comparison is done using robust ranking statistics, by testing if two images share the same subset of $k$ maximally activated representation components.
This test is used to decide if two unlabelled images belong to the same (new) class or not, generating a set of noisy pairwise pseudo-labels.
The pseudo-labels are then used to learn a similarity function for the unlabelled images.

The third idea is, after bootstrapping the representation, to optimise the model by minimizing a joint objective function, containing terms for both the labelled and unlabelled subsets. To do this, we use respectively the given labels and the generated pseudo-labels, thus avoiding the forgetting issue~\cite{mccloskey1989catastrophic} that may arise with a sequential approach.
A further boost is obtained by incorporating incremental learning of the discovered classes in the classification task, which allows information to flow between the labelled and unlabelled images.

However, this approach still requires knowing the number of new categories in the unlabelled data, which is not a realistic assumption in many applications.
We propose a method to estimate the number of classes in the unlabelled data
which also transfers knowledge from the set of known classes.
The idea is to use part of the known classes as a probe set, adding them to the unlabelled set pretending that part of them are unlabelled, and then running the clustering algorithm described above on the extended unlabelled dataset.
This allows us to cross-validate the number of new classes, according to the clustering accuracy on the probe set as well as a cluster quality index on the unlabelled set, resulting in a reliable estimate of the true number of unlabelled classes.

In addition, AutoNovel can be used for unsupervised image clustering by simply removing the requirement of labelled data, resulting in a simplified version of our method, achieving the state-of-the-art results on image clustering.

We evaluate AutoNovel on several public benchmarks, outperforming by a large margin all existing techniques that can be applied to novel category discovery, demonstrating the effectiveness of our approach. We also evaluate our category number estimation method, showing reliable estimation of the number of categories in the unlabelled data.
We also demonstrate the effectiveness of our method by using it for unsupervised image clustering, achieving state-of-the-art clustering results.

We have presented preliminary results of this work in~\cite{han20automatically}, and this paper extends them in several aspects. First, we include a solution to handle the case of an unknown number of categories in unlabelled data, which we initially introduced in~\cite{han2019learning}.
Second, we study different alternatives to ranking statistics for generating pairwise pseudo labels and compare their effectiveness.
Third, we expand the experiments and study transferring representations from ImageNet-pretrained models to new domains.
Fourth, we test the effectiveness of different self-supervised learning methods when used as a component of our method.
Fifth, we show that our method can also be used for unsupervised clustering, achieving the state-of-the-art results on public benchmarks.

The code reproducing our experiments can be found at \url{http://www.robots.ox.ac.uk/~vgg/research/auto_novel}.

\section{Related work}\label{sec:related_work}

Our work draws inspiration from semi-supervised learning, transfer learning,  clustering, and  zero-shot learning.
We review below the most relevant contributions.

In semi-supervised learning (SSL)~\cite{chapelle2006semi}, a partially labelled training dataset is given and the objective is to learn a model that can propagate the labels from the labelled data to unlabelled data.
Most SSL methods focus on the classification task where, usually, both labelled and unlabelled points belong to the same set of classes.
On the contrary, our goal is to handle the case where the unlabelled data classes differ from the labelled data.
\hbox{}\cite{oliver2018realistic} summarizes the state-of-the-art SSL methods.
Among them, the consistency-based methods appear to be the most effective.
\hbox{}\cite{rasmus2015semi} proposes a ladder network which is trained on both labelled and unlabelled data using a reconstruction loss.
\hbox{}\cite{laine2016temporal} simplifies this ladder network by enforcing prediction consistency between a data point and its augmented counterpart.
As an alternative to data augmentation, they also consider a regularization method based on the exponential moving average (EMA) of the predictions.
This idea is further improved by~\cite{tarvainen2017mean}:
instead of using the EMA of predictions, they propose to maintain the EMA of model parameters.
The consistency is then measured between the predictions of the current model (student) and the predictions of the EMA model (teacher).
More recently (and closer to our work) practitioners have also combined SSL with self-supervision\cite{rebuffi2019semi,48416} to leverage datasets with very few annotations.
\rev{Finally, FixMatch \cite{sohn2020fixmatch} also uses pseudo-labels extracted from the most confident images reaching state-of-the-art results in SSL benchmarks.
However, they use pseudo-labels as soft targets for the cross-entropy loss, while we use a binary score to evaluate similarity of sample pairs within a mini-batch.}
 
Transfer learning~\cite{Pan10transfer,weiss2016asurvey,tan2018asurvey} is an effective way to reduce the amount of data annotations required by pre-training the model on a different dataset.
In image classification, for example, it is customary to start from a model pre-trained on the ImageNet~\cite{deng09imagnet} dataset.
In most transfer learning settings, however, both the source data and the target data are fully annotated.
In contrast, our goal is to transfer information from a labelled dataset to an unlabelled one.

Many classic (e.g.,~\cite{Aggarwal13cluster,MackQueen67_Kmeans,Comaniciu02meanshift,ng2001onspectral}) and deep learning (e.g.,~\cite{Xie16_DEC,Chang_2017_ICCV,Dizaji2017deepclustering,Yang17towards,yang2016joint,Hsu18_L2C,Hsu19_MCL}) clustering methods have been proposed to automatically partition an unlabelled data collection into different classes.
However, this task is usually ill-posed as there are multiple, equally valid criteria to partition most datasets.
We address this challenge by learning the appropriate criterion by using a labelled dataset, narrowing down what constitutes a proper class.
We call this setting ``transfer clustering''.

To the best of our knowledge, the work most related to ours are~\cite{Hsu18_L2C, Hsu19_MCL, han2019learning}. In \cite{han2019learning}, the authors also consider discovering new classes as a transfer clustering problem. They first learn a data embedding by using metric learning on the labelled data, and then fine-tune the embedding and learn the cluster assignments on the unlabelled data.
In \cite{Hsu18_L2C, Hsu19_MCL}, the authors introduce KCL and MCL clustering methods. In both, a similarity prediction network (SPN), also used in \cite{hsu2016deep}, is first trained on a labelled dataset.
Afterwards, the pre-trained SPN is used to provide binary pseudo labels for training the main model on an unlabelled dataset.
The overall pipelines of the two methods are similar, but the losses differ:
KCL uses a contrastive loss based on Kullback-Leibler divergence, which is equivalent to the BCE used in this paper (eq.~(\ref{e:bce})), and MCL uses the Meta Classification Likelihood loss.
Zero-shot learning (ZSL)~\cite{Xian2018zsl,fu2018recent} can also be used to  recognize new classes.
However, differently from our work, ZSL also requires additional side information (e.g., class attributes) in addition to the raw images.

Finally, other works~\cite{Dean2013fast,Yagnik2011thepower} discuss the application of ranking statistics to measuring the similarity of vectors; however, to the best of our knowledge, we are the first to apply ranking statistics to the task of novel category discovery using deep neural networks.
\section{Method}\label{sec:method}
Given an \emph{unlabelled} dataset \rev{$D^u=\{x_i^u; i=1,\dots,M\}$} of images $x_i^u \in \mathbb{R}^{3\times H\times W}$, our goal is to automatically cluster them into a number of classes $C^u$.
We also assume to have a second \emph{labelled} image dataset \rev{$D^l=\{(x_i^l,y_i^l); i=1,\dots,N\}$} where $y_i^l\in \{ 1,\dots, C^l \}$ is the class label for image $x_i^l$, where the set of $C^l$ labelled classes is disjoint from the set of $C^u$ unlabelled ones.
While the statistics of $D^l$ and $D^u$ differ, we hypothesize that a general notion of what constitutes a ``good class'' can be extracted from $D^l$ and that the latter can be used to better cluster $D^u$.

We approach the problem by learning an image representation $\Phi : x \mapsto \Phi(x) \in \mathbb{R}^d$ in the form of a CNN\@.
The goal of the representation is to recognize the known classes and to discover the new ones.
In order to learn this representation, we propose AutoNovel, a method that combines three ideas detailed in the next three sections.

\subsection{Self-supervised learning}\label{s:slefsup}

Given that we have a certain number of labelled images $D^l$ at our disposal, the obvious idea is to use these labels to bootstrap the representation $\Phi$ by minimizing a standard supervised objective such as the cross-entropy loss.
However, experiments show that this causes the representation to overly-specialize for the classes in $D^l$, providing a poor representation of the new classes in $D^u$.

Thus we resist the temptation of using the labels right away and instead use a self-supervised learning method to bootstrap the representation $\Phi$.
Self-supervised learning has been shown~\cite{kolesnikov2019revisiting,gidaris2018unsupervised} to produce robust low-level features, especially for the first few layers of typical CNNs.
It has the benefit that no data annotations are needed, and thus it can be applied to both labelled and unlabelled images during training.
In this way, we achieve the key benefit of ensuring that the representation is initialized without being biased towards the labelled data.

In detail, we first pre-train our model $\Phi$ with self-supervision on the union of $D^l$ and $D^u$ (ignoring all labels).
We use the RotNet~\cite{gidaris2018unsupervised} approach%
\footnote{
We present to the network $\Phi$ randomly-rotated versions $R(x)$ of each image and task it with predicting $R$.
The problem is formulated as a 4-way classification of the rotation angle, with angle in $\{0^\circ,90^\circ,180^\circ,270^\circ\}$.
The model $\eta \circ \Phi(R(x))$ is terminated by a single linear  layer $\eta$ with 4 outputs each scoring an hypothesis.
The parameters of $\eta$ and $\Phi$ are optimized by minimizing the cross-entropy loss on the rotation prediction.} as our default choice
due to its simplicity and efficacy, but any self-supervised method could be used instead.
In our experiments, we also experimented with other self-supervised learning methods such as SimCLR~\cite{chen2020simple} and MoCo~\cite{he2020moco}. Interestingly, we found these alternatives perform less effectively than RotNet in our setting, though they have shown better performance for fully supervised downstream tasks such as recognition and detection.
We then extend the pre-trained network $\Phi$ with a classification head $\eta^l:\mathbb{R}^d \rightarrow \mathbb{R}^{C^l}$ implemented as a single linear layer followed by a softmax layer.
The function $\eta^l \circ \Phi$ is fine-tuned on the labelled dataset $D^l$ in order to learn a classifier for the $C^l$ known classes, this time using the labels $y_i$ and optimizing the standard cross-entropy (CE) loss:
\begin{equation}\label{e:ce}
L_\text{CE} =
- \frac{1}{N}
\sum_{i=1}^N
\log [\eta^l\circ z^l_i]_{y_i}
\end{equation}
where $z^l_i=\Phi(x^l_i) \in \mathbb{R}^{d}$ is the representation of image $x^l_i$.
Only $\eta^l$ and the last macro-block of $\Phi$ (see \cref{sec:exp} for details) are updated in order to avoid overfitting the representation to the labelled data.

\subsection{Transfer learning via ranking statistics}\label{s:rank}

Once the representation $\Phi$ and the classifier $\eta^l$ have been trained, we are ready to look for the new classes in $D^u$.
Since the classes in $D^u$ are unknown, we represent them by defining a relation among pairs of unlabelled images $(x_i^u, x_j^u)$.
The idea is that similar images should belong to the same (new) class, which we denote by the symbol $s_{ij}=1$, while dissimilar ones should not, which we denote by $s_{ij}=0$. \rev{A similar idea has been applied in the literature (e.g., \cite{Hsu18_L2C,Hsu19_MCL,Chang_2018_DAC,rebuffi2020lsdc}).} 
The problem is then to obtain the labels $s_{ij}$.

Our assumption is that the new classes will have some degree of visual similarity with the known ones.
Hence, the learned representation should be applicable to old and new classes equally well.
As a consequence, we expect the descriptors $z^u_i=\Phi(x^u_i)$ and $z^u_j=\Phi(x^u_j)$ of two images $x_i^u$, $x_j^u$ from the new classes to be close if they are from the same (new) class, and to be distinct otherwise.
The way this is done is explained in the next section.

\subsubsection{Ranking statistics}
\label{sec:method:ranking}
\begin{figure}
\centering
\includegraphics[width=\linewidth]{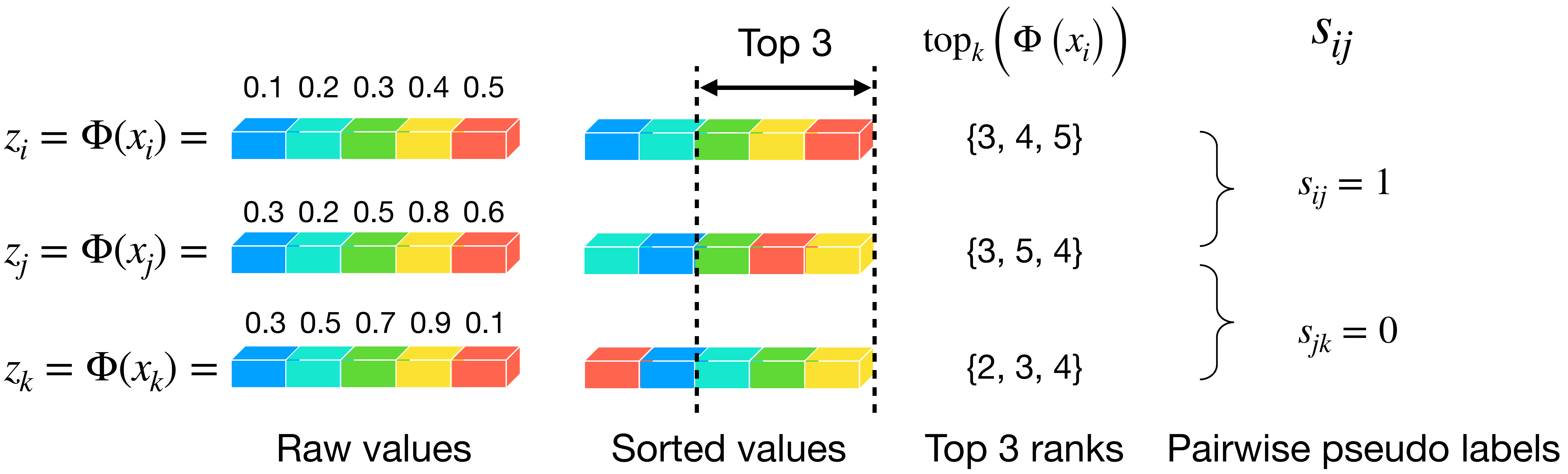}
\caption{Ranking statistics. In this example, we consider top-3 ranks. As the top-3 ranks of $z_i$ and $z_j$ are the same, $s_{ij}=1$. While the top-3 ranks of $z_j$ and $z_k$ are the different, so $s_{jk}=0$.}\label{fig:ranking}
\end{figure}

Rather than comparing vectors $z^u_i$ and $z^u_j$ directly (e.g., by a scalar product), we use a more robust ranking statistics.
Specifically, we rank the values in vector $z^u_i$ by magnitude.
Then, if the rankings obtained for two unlabelled images $x^u_i$ and $x^u_j$  are the same, they are very likely to belong to the same (new) class, so we set $s_{ij} = 1$.
Otherwise, we set $s_{ij} = 0$.
In practice, it is too strict to require the two rankings to be identical if the dimension of $z^u_i$ is high (otherwise we may end up with $s_{ij}=0$ for all pairs $(i,j), i\neq j$).
Therefore, we relax this requirement by only testing if the \emph{sets} of the top-$k$ ranked dimensions are the same (we use $k=5$ in our experiments), i.e.:
\begin{equation}\label{e:ranking}
s_{ij} =
\mathbbm{1}
\left \{
\operatorname{top}_k(\Phi(x^u_i)) =
\operatorname{top}_k(\Phi(x^u_j))
\right \},
\end{equation}
where $\operatorname{top}_k : \mathbb{R}^d \rightarrow \mathcal{P}(\{1,\dots,d\})$ associates to a vector $z$ the subset of indices $\{1,\dots,d\}$ of its top-$k$ elements. \Cref{fig:ranking} shows an example of using ranking statistics to obtain pairwise pseudo labels.

Once the values $s_{ij}$ have been obtained, we use them as pseudo-labels to train a comparison function for the unlabelled data.
In order to do this, we apply a new head $\eta^u :\mathbb{R}^d \rightarrow\mathbb{R}^{C^u}$ to the image representation $z^u_i=\Phi(x^u_i)$ to extract a new descriptor vector $\eta^u(z^u_i)$ optimized for the unlabelled data.
As in \cref{s:slefsup}, the head is composed of a linear layer followed by a softmax.
Then, the inner product $\eta^u(z^u_i)^\top \eta^u(z^u_j)$ is used as a score for whether images $x_i^u$ and $x_j^u$ belong to the same class or not.
Note that $\eta^u(z^u_i)$ is a normalized vector due to the softmax layer in $\eta^u$.
This descriptor is trained by optimizing the \emph{binary cross-entropy} (BCE) loss:
\begin{equation}\label{e:bce}
\begin{aligned}
L_\text{BCE} =
& - \frac{1}{M^2}
\sum_{i=1}^M
\sum_{j=1}^M
[
s_{i j}
\log \eta^u(z^u_i)^\top\eta^u(z^u_j) \\
& +
(1 - s_{i j})
\log (1 - \eta^u(z^u_i)^\top\eta^u(z^u_j))
].
\end{aligned}
\end{equation}

Furthermore, we structure $\eta^u$ in a particular manner:
\rev{We set its output dimension to be equal to the number of new classes $C^u$, which is a common practice for clustering in the literature (e.g.,~\cite{Hsu18_L2C,Hsu19_MCL,Chang_2018_DAC,rebuffi2020lsdc}).}
In this manner, we can use the index of the maximum element of each vector $\hat y_i^u = \operatorname{argmax}_y [\eta^u\circ\Phi(x_i^u)]_y$ as prediction $\hat y_i^u$ for the class of image $x_i^u$ (as opposed to assigning labels via a clustering method such as $k$-means).

\smallskip\noindent\textbf{Alternatives to ranking statistics.}
While we adopt ranking statistics to obtain the pairwise pseudo labels for unlabelled data, there exist many other options such as $k$-means, cosine similarity, and nearest neighbor.
In the experiment, we evaluate applying $k$-means on the unlabelled data in each mini-batch and use the resulting cluster assignments to generate pairwise pseudo labels.
\rev{We can also compute cosine similarity between vectors and generate binary pseudo-labels based on a predefined threshold $\tau$. Another natural way to generate binary pseudo-labels is using the  mutual nearest neighbor criteria, for which we follow~\cite{Sarfraz19finch} and define} 
\begin{equation}\label{e:mutual_nn}
s_{ij}=\left\{\begin{array}{ll}
1 & \text { if } j=\kappa_{i}^{1} \text { or } \kappa_{j}^{1}=i \text { or } \kappa_{i}^{1}=\kappa_{j}^{1} \\
0 & \text { otherwise }
\end{array}\right.,
\end{equation}
where $\kappa_{i}^{1}$ denotes the nearest neighbor of image $i$ in the mini-batch.
As will be shown in the experiments, these alternatives are less effective than using ranking statistics.

\smallskip\noindent\textbf{Discussion.}
There are several reasons why we believe the ranking statistics do well.
First, the statistics focus on the top-$k$ most active feature components for each image.
Intuitively, the magnitude of these components reflects the degree to which they are discriminative for the object in the image.
Thus, our ranking statistics only considers the most salient feature components when comparing images, while ignoring noisy components with small values.

Second, other similarity measures like cosine similarity, which use the whole feature vectors for comparison in a high-dimensional vector space, can potentially suffer from the problem of distance concentration~\cite{kab11onthedistance}.
The distance concentration is the counter-intuitive phenomenon that, as the data dimensionality increases, all pairwise distances between points may converge to the same value, which is not desired in our case.

Third, we further relax the comparison by not requiring the order of the top-$k$ ranks to be identical.
Instead, we only check the \emph{sets} of the top-$k$ ranks.
This further makes the pairwise comparisons robust to slight discrepancies among the most discriminative feature components.

\subsection{Joint training on labelled and unlabelled data}\label{s:joint}

We now have two losses that involve the representation $\Phi$: the CE loss $L_\text{CE}$ for the labelled data $D^l$ and the pairwise BCE loss $L_\text{BCE}$ for the unlabelled data $D^u$.
They both share the same image embedding $\Phi$.
This embedding can be trained sequentially, first on the labelled data, and then on the unlabelled data using the pseudo-labels obtained above.
However, in this way the model will very likely forget the knowledge learned from the labelled data, which is known as \emph{catastrophic forgetting} in incremental learning~\cite{rebuffi2017icarl,lopez2017gradient,shmelkov2017incremental,aljundi2018memory}.

\begin{figure}
\centering
\includegraphics[width=0.9\linewidth]{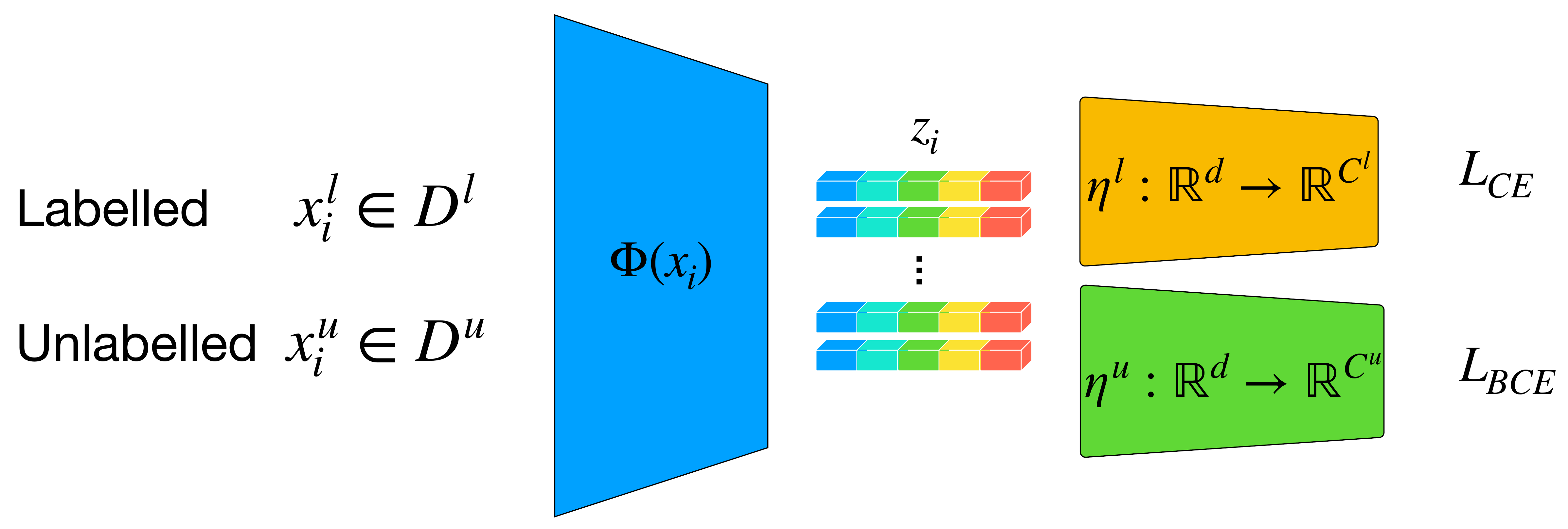}
\caption{Joint learning on labelled and unlabelled data.}\label{fig:arch}
\end{figure}

Instead, we jointly fine-tune our model using both losses at the same time.
Note that most of the model $\Phi$ is frozen; we only fine-tune the last macro-block of $\Phi$ together with the two heads $\eta^u$ and $\eta^l$.
\Cref{fig:arch} demonstrates the overall architecture for joint learning.
Importantly, as we fine-tune the model, the labels $s_{ij}$ are changing at every epoch as  the embedding $\eta^l$ is updated.
This in turn affects the ranking statistics used to determine the labels $s_{ij}$ as explained in~\cref{s:rank}.
This leads to a ``moving target'' phenomenon that can introduce some instability in learning the model.
This potential issue is addressed in the next section.

\subsection{Enforcing predictions to be consistent}\label{s:consistency}

In addition to the CE and BCE losses, we also introduce a consistency regularization term, which is used for both labelled and unlabelled data.
In semi-supervised learning~\cite{oliver2018realistic,tarvainen2017mean,laine2016temporal}, the idea of consistency  is that the class predictions on an image $x$ and on a randomly-transformed counterpart \rev{$t(x)$} (for example an image rotation) should be the same.
In our case, as will be shown in the experiments, consistency is important to obtain good performance.
One reason is that, as noted above, the pairwise pseudo-labels for the unlabelled data are subject to change on the fly during training.
Indeed, for an image $x^u_i$ and a randomly-transformed counterpart \rev{$t(x^u_i)$}, if we do not enforce consistency, we can have \rev{$\operatorname{top}_k(\Phi(x^u_i)) \neq \operatorname{top}_k(\Phi(t(x^u_i)))$}.
According to~\cref{e:ranking} defining $s_{ij}$, this could result in different $s_{ij}$ for $(x^u_i, x^u_j)$ depending on the data augmentation applied to the images.
This variability of the ranking labels for a given pair could then confuse the training of the embedding.

Following the common practice in semi-supervised learning, we use the \emph{Mean Squared Error} (MSE) as the consistency cost.
This is given by:
\begin{equation}
    L_\text{MSE} =
   \frac{1}{N} \sum_{i=1}^{N}(\eta^l(z^l_i)-\eta^l(\hat{z}^l_i))^{2}
 + \frac{1}{M} \sum_{i=1}^{M}(\eta^u(z^u_i)-\eta^u(\hat{z}^u_i))^{2},
\end{equation}
where $\hat{z}$ is the representation of \rev{$t(x)$}.

The overall loss of our model can then be written as
\begin{equation}\label{eq:loss}
    L = L_\text{CE} + L_\text{BCE} + \omega(r)L_\text{MSE},
\end{equation}
where the coefficient $\omega(r)$ is a ramp-up function.
This is widely used in semi-supervised learning~\cite{laine2016temporal,tarvainen2017mean}.
Following~\cite{laine2016temporal,tarvainen2017mean}, we use the sigmoid-shaped function
$
\omega(r) = \lambda e^{-5(1-\frac{r}{T})^{2}}
$,
where $r$ is current time step and $T$ is the ramp-up length and $\lambda \in \mathbb{R}_+$.
\rev{As opposed to contrastive learning~\cite{chen2020simple} this loss is only minimising the distance between positive pairs of samples within a mini-batch.}

\subsection{Incremental learning scheme}\label{s:incremental}

We also explore a setting analogous to incremental learning.
In this approach, after tuning on the labelled set (end of~\cref{s:slefsup}), we extend the head $\eta^l$ to $C^{u}$ new classes, so that $\eta^l:\mathbb{R}^d \rightarrow \mathbb{R}^{C^{l}+C^{u}}$.
The head parameters for the new classes are initialized randomly.
The model is then trained using the same loss~\cref{eq:loss}, but the cross-entropy part of the loss is evaluated on both labelled and unlabelled data $D^l$ and $D^u$.
Since the cross-entropy requires labels, for the unlabelled data we use the \emph{pseudo-labels} $\hat y_i^u$, which are generated on-the-fly from the head $\eta^u$ at each forward pass.

The advantage is that this approach \emph{increments} $\eta^l$ to discriminate both old and new classes, which is often desirable in real applications.
It also creates a feedback loop that causes the features $z^u_i$ to be refined, which in turn generates better pseudo-labels $\hat y_i^u$ for $D^u$ from the head $\eta^u$.
In this manner, further improvements can be obtained by this cycle of positive interactions between the two heads during training.

\subsection{Unsupervised clustering}\label{s:clustering}
Rather than working on the task of novel category discovery, AutoNovel can also be used for standard (unsupervised) clustering by simply removing the use of labelled data (thus dropping step two of the method), obtaining a two-step approach.
In the first step, we pre-train our model with self-supervised learning as before.
In the second step, we finetune the last macro block and the linear layer for clustering, using the ranking statistics to provide pseudo pairwise labels.
This way, our method can simultaneously learn the feature embedding for clustering as well as the cluster assignments.
The training loss in~\cref{eq:loss} becomes:
\begin{equation}\label{eq:cluster_loss}
    L = L_\text{BCE} + \omega(r)L_\text{MSE}.
\end{equation}

To our knowledge, Deep Adaptive
Clustering (DAC)~\cite{Chang_2018_DAC} is the deep clustering method most related to this approach, in the sense that DAC also reduces clustering as a binary classification problem.
However, our approach differs from DAC in several aspects.
First, our method uses ranking statistics to generate pairwise pseudo labels instead of the cosine similarity (though any pairwise labeling method could also be used in our approach).
Second, our method optimizes the standard binary cross-entropy loss with a consistency constraint, while DAC optimizes a Bhattacharyya distance with an ad-hoc sample number penalty term.
Third, our method incorporates self-supervised learning for pretraining lower level features and only needs to train the last macro-block and the linear layers of the model, which is less likely to suffer from overfitting.
By comparison, DAC updates all parameters of the model.
As will be seen in the experiment, our method significantly outperforms DAC as well as the recent state-of-the-art method IIC~\cite{ji2019invariant} in several benchmarks.
Note that the main objective of this work is novel category discovery rather than clustering, and here we only show that our method can be easily adopted for the problem of clustering achieving superior performance than exiting alternatives.

\section{Estimating the number of classes}\label{s:k}

\begin{algorithm}[t]
\caption{Estimating the number of classes}\label{alg:unknown_k}
\begin{algorithmic}[1]
\State \textbf{Preparation:}
\State Split the probe set $D^l_r$ into $D^l_{ra}$ and $D^l_{rv}$.
\State Extract features of $D^l_r$ and $D^u$ using $\Phi$.
\State \textbf{Main loop:}
\For{$0\leq C^u_i \leq C^u_\text{max}$}
  \State
  Run $k$-means on $D^l_r \cup D^u$ assuming $C^{lu}_r=C^l_r+C^u_i$ classes in semi-supervised mode (i.e.~forcing data in $D^l_{ra}$ to map to the ground-truth class labels).
  \State Compute ACC for $D^l_{rv}$ and CVI for $D^u$.
\EndFor
\State \textbf{Obtain optimal:}
\State Let $C^{u*}_a$ be the value of $C^u_i$ that maximise ACC for $D^l_{rv}$ and $C^{u*}_v$ be the value that maximise CVI for $D^u$ and let $\hat C^u = (C^{u*}_a + C^{u*}_v)/2$. Run semi-supervised $k$-means on $D^l_r \cup D^u$ again assuming $C^l_r+\hat C^u$ classes.
\State \textbf{Remove outliers:}
\State Look at the resulting clusters in $D^u$ and drop any that has a mass less than $\tau$ of the largest cluster. Output the number of remaining clusters.
\end{algorithmic} 
\end{algorithm}

So far, we have assumed that the number of classes $C^u$ in the unlabelled data is known, but this is usually not the case in real applications.
Here we propose a new approach to estimate the number of classes in the unlabelled data by making use of labelled probe classes.
The probe classes are combined with the unlabelled data and the resulting set is clustered using $k$-means multiple times, varying the number of classes.
The resulting clusters are then examined by computing two quality indices, one of which checks how well the probe classes, for which ground truth data is available, have been identified.
The number of categories is then estimated to be the one that maximizes these quality indices.

\begin{figure}
\centering
\includegraphics[width=0.75\linewidth]{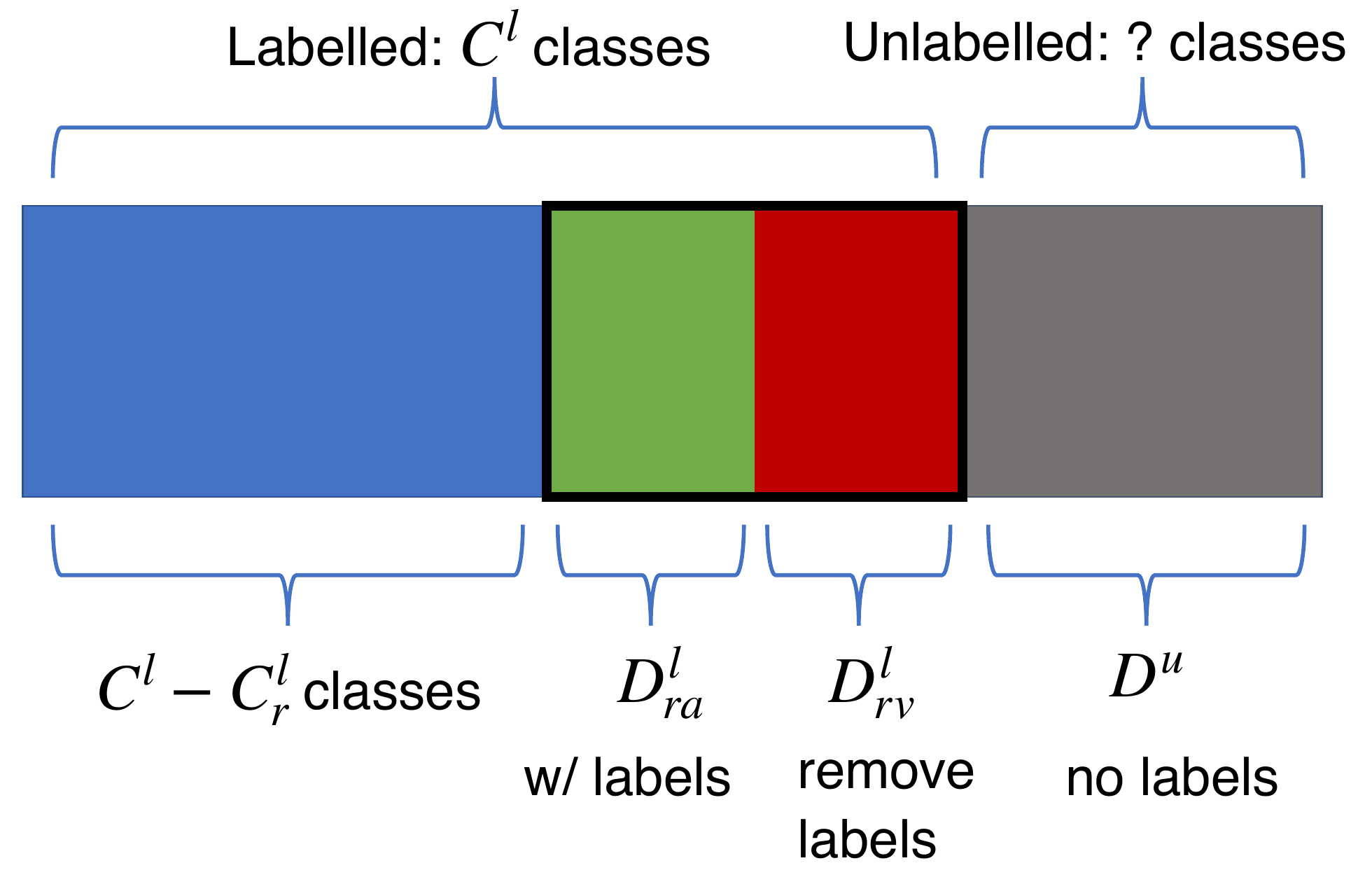}
\caption{Data split for category number estimation.}\label{fig:k_est_split}
\end{figure}

In more details, we first split the $C^l$ known classes into a probe subset $D^l_r$ of $C^l_r$ classes and a training subset $D^l \setminus D^l_r$ containing the remaining $C^l-C^l_r$ classes. These $C^l-C^l_r$ classes are used for supervised feature representation learning, while the $C^l_r$ probe classes are combined with the unlabelled data for class number estimation. We then further split the $C^l_r$ probe classes into a subset $D^l_{ra}$ of $C^l_{ra}$ classes and a subset $D^l_{rv}$ of $C^l_{rv}$ classes (e.g.,  $C^l_{ra} : C^l_{rv} = 4:1$), which we call anchor probe set and validation probe set respectively (see~\cref{fig:k_est_split}).
We then run a constrained (semi-supervised) $k$-means on $D^l_r \cup D^u$ to estimate the number of classes in $D^u$. Namely, during $k$-means, we force images in the anchor probe set $D^l_{ra}$ to map to clusters following their ground-truth labels, while images in the validation probe set $D^l_{rv}$  are considered as additional ``unlabelled'' data.
We launch this constrained $k$-means multiple times by sweeping the number of total categories $C_r^{lu}$ in $D^l_r \cup D^u$, and measure the constrained clustering quality on  $D^l_r \cup D^u$.
We consider two quality indices, given below, for each value of $C_r^{lu}$.
The first measures the cluster quality in the labelled validation probe set $D^l_{rv}$, whereas the second measures the quality in the unlabelled data $\mathcal{D}^u$.
Each index is used to determine an optimal number of classes and the results are averaged.
Finally, $k$-means is run one last time with this value as the number of classes and any outlier clusters in $D^u$, defined as containing less than $\tau$ (e.g., $\tau = 1\%$) the mass of the largest clusters, are dropped.
The details are given in \rev{method summary~\ref{alg:unknown_k}}.

\smallskip\noindent\textbf{Cluster quality indices}.
We measure our clustering for class number estimation with two indices.
The first index is the \emph{average clustering accuracy} (ACC), which is applicable to the $C^l_{rv}$ labelled classes in the validation probe set $D^l_{rv}$ and is given by
\begin{equation}\label{e:acc}
  \operatornamewithlimits{max}_{g \in \operatorname{Sym}(C^l_{rv})}
  \frac{1}{N} \sum_{i=1}^{N} \mathbf{1}\left\{\bar{y}_{i}=g\left(y_{i}\right)\right\},
\end{equation}
where $\overline{y}_{i}$ and ${y}_{i}$ denote the ground-truth label and clustering assignment for each data point $x_i\in D^l_{rv}$ and $\operatorname{Sym}(C^l_{rv})$ is the group of permutations of $C^l_{rv}$ elements (this discounts the fact that the cluster indices may not be in the same order as the ground-truth labels).
Permutations are optimized using the Hungarian algorithm~\cite{kuhn1955hungarian}.

The other index is a \emph{cluster validity index} (CVI)~\cite{Arbelaitz12cluster} which, by capturing notions such as  intra-cluster cohesion vs inter-cluster separation, is applicable to the unlabelled data $D^u$.
There are several CVI metrics, such as Silhouette~\cite{Rousseeuw87Silhouettes}, Dunn~\cite{Dunn74dunn}, Davies–Bouldin~\cite{Davies79pami}, and Calinski-Harabasz~\cite{Cali74cluster}; while no metric is uniformly the best, the Silhouette index generally works well~\cite{Arbelaitz12cluster,Bezdek98cluster}, and we found it to be a good choice for our case too.
This index is given by
\begin{equation}\label{e:silhouette}
\sum_{x\in\mathcal{D}^u} \frac{b(x) - a(x)}{\max\{a(x),b(x)\}},
\end{equation}
where $x$ is a data sample, $a(x)$ is the average distance between $x$ and all other data samples within the same cluster, and $b(x)$ is the smallest average distance of $x$ to all points in any other cluster (of which $x$ is not a member).
\section{Experiments}\label{sec:exp}

\subsection{Data and implementation details}\label{sec:exp:imp}

We evaluate AutoNovel on a variety of standard benchmark datasets:
CIFAR10~\cite{Krizhevsky09cifar}, CIFAR100~\cite{Krizhevsky09cifar}, SVHN~\cite{Netzer2011svhn}, OmniGlot~\cite{Lake15omnniglot}, and ImageNet~\cite{deng09imagnet}.  Following~\cite{han2019learning}, we split these to have
 5/20/5/654/30 classes respectively in the unlabelled set. The splits are summarized in~\cref{tab:datasplit}.
In addition, for OmniGlot and ImageNet we use 20 and 3 different splits respectively, as in~\cite{han2019learning}, and report average clustering accuracy (as defined in~\cref{e:acc}) on the unlabelled data.
\rev{While we follow standard practice to split the datasets we note here that most of the time, the number of unlabelled classes is under a hundred. This is a potential limitation of clustering which still proves to be very difficult for classification over thousands of categories \cite{vangansbeke2020scan}.}

\begin{table}[ht]
\centering
\footnotesize
\caption{Data splits in the experiments.}\label{tab:datasplit}
\begin{tabular}{lcc}
\toprule
& labelled classes & unlabelled classes \\
\midrule
CIFAR10 & 5 & 5 \\
CIFAR100 & 80 &  20 \\
SVHN & 5  & 5 \\
OmniGlot & 964  & 659 \\
ImageNet & 882 & 118 \\
\bottomrule
\end{tabular}
\end{table}

We use the ResNet-18~\cite{he2016deep} architecture, except for OmniGlot for which we use a VGG-like network~\cite{simonyan15vgg} with six layers to make our setting directly comparable to prior work.
We use SGD with momentum~\cite{sutskever2013importance} as the optimizer for all but the OmniGlot dataset, for which we use Adam~\cite{kingma2014adam}.
For all experiments we use a batch size of 128 and $k=5$ which we found work consistently well across datasets.

In the first self-supervised training step, \rev{unless} otherwise mentioned, we train our model with the pretext task of rotation predictions (i.e., a four-class classification: $0^{\circ}$, $90^{\circ}$, $180^{\circ}$, and $270^{\circ}$) for 200 epochs and a step-wise decaying learning rate starting from 0.1 and divided by 5 at epochs 60, 120, and 160.

In the second step of our framework (i.e., supervised training using labelled data), we fine-tune our model on the labelled set for 100 epochs and a step-wise decaying learning rate starting from 0.1 and halved every 10 epochs.
From this step onward we fix the first three convolutional blocks of the model, and fine-tune the last convolutional block together with the linear classifier.

Finally, in the last joint training step, we fine-tune our model for 200/100/90 epochs for \{CIFAR10, CIFAR100, SVHN\}/OmniGlot/ImageNet, which is randomly sampled from the merged set of both labelled and unlabelled data.
The initial learning rate is set to 0.1 for all datasets, and is decayed with a factor of 10 at the 170th/\{30th, 60th\} epoch for \{CIFAR10, CIFAR100, SVHN\}/ImageNet.
The learning rate of 0.01 is kept fixed for OmniGlot.
For the consistency regularization term, we use the ramp-up function as described in~\cref{s:consistency} with $\lambda = \{5.0, 50.0, 50.0, 100.0, 10.0\}$, and $T = \{50, 150, 80, 1, 50\}$ for CIFAR10, CIFAR100, SVHN, OmniGlot, and ImageNet respectively.

In the incremental learning setting, all previous hyper parameters remain the same for our method.
We only add a ramp-up on the cross entropy loss on unlabelled data.
The ramp-up length is the same as the one used for eq.~(4) and we use for all experiments a coefficient of 0.05.
For all other methods we train the classifier for 150 epochs with SGD with momentum and learning rate of 0.1 divided by 10 at epoch 50.

\rev{For hyper-parameter tuning, we create a probe validation set from the labelled data by dropping the labels of a few classes. We then tune the hyper-parameters based on the ACC on this probe validation set. We construct the probe validation set to have the same number of classes as the actual unlabelled set. For example, for CIFAR100, we split the 80 labelled classes into a 60-class labelled subset and a 20-class probe validation set. We then tune the hyper-parameters based on the novel category discovery performance on the probe validation set. For CIFAR10 and SVHN, due to the small number of labelled classes, we only take 2 classes from the labelled data to construct the probe validation set.}

We implement our method using PyTorch 1.1.0 and run experiments on NVIDIA Tesla M40 GPUs. Following \cite{han2019learning}, our results are averaged over 10 runs for all datasets, except ImageNet, for which the results are averaged over the three 30-class subsets.  In general, we found the results are stable. Our code is publicly available at \url{http://www.robots.ox.ac.uk/~vgg/research/auto_novel}.

\begin{table}[t]
\centering
\caption{Ablation study of AutoNovel. ``MSE'' means MSE consistency constraint; ``CE'' means cross entropy loss for training on labeled data; ``BCE'' means binary cross entropy loss for training on unlabeled data; ``S.S.'' means self-supervision; \rev{``I.L.'' means incremental learning.} \rev{The evaluation metric is the ACC.}}\label{tab:ablation_unlabelled}
\begin{tabular}[c]{lccc}
\toprule
          & CIFAR-10     & CIFAR-100     & SVHN \\
\midrule
Ours w/o MSE & 82.6$\pm$12.0\% & 61.8$\pm$3.6\% & 61.3$\pm$1.9\%   \\
Ours w/o CE & 84.7$\pm$4.4\%  & 58.4$\pm$2.7\% & 59.7$\pm$6.6\%  \\
Ours w/o BCE & 26.2$\pm$2.0\%  & 6.6$\pm$0.7\% & 24.5$\pm$0.5\% \\
Ours w/o S.S.& 89.4$\pm$1.4\%  & 67.4$\pm$2.0\% & 72.9$\pm$5.0\% \\\midrule
Ours full   & \textbf{90.4$\pm$0.5}\% & \textbf{73.2$\pm$2.1}\%  & \textbf{95.0$\pm$0.2}\% \\
Ours w/ I.L.  & \textbf{91.7$\pm$0.9}\% & \textbf{75.2$\pm$4.2}\% & \textbf{95.2$\pm$0.3}\% \\
\bottomrule
\end{tabular}\hfill%

\end{table}

\subsection{Ablation study}

We validate the effectiveness of the components of AutoNovel by ablating them and measuring the resulting ACC on the unlabelled data.
Note that, since the evaluation is restricted to the unlabelled data, we are solving a clustering problem. The same unlabelled data points are used for both training and testing, except that data augmentation (i.e.~image transformations) is not applied when computing the cluster assignments.
As can be seen in~\cref{tab:ablation_unlabelled}, all components have a significant effect as removing any of them causes the performance to drop substantially.
Among them, the BCE loss is by far the most important one, since removing it results in a dramatic drop of 40--60\% absolute ACC points.
For example, the full method has ACC $90.4\%$ on CIFAR10, while removing BCE causes the ACC to drop to $26.2\%$.
This shows that that our rank-based embedding comparison can indeed generate reliable pairwise pseudo labels for the BCE loss.
Without consistency, cross entropy, or self-supervision, the performance drops by a more modest but still significant $7.8\%$, $5.7\%$ and $1.0\%$ absolute ACC points, respectively, for CIFAR10.
It means that the consistency term plays a role as important as the cross-entropy term by preventing the ``moving target'' phenomenon described in~\cref{s:consistency}.
Finally, by incorporating the discovered classes in the classification task, we get a further boost of 1.3\%, 2.0\% and 0.2\% points on CIFAR10, CIFAR100 and SVHN respectively.

\begin{figure}
    \centering
    \includegraphics[width=0.8\linewidth]{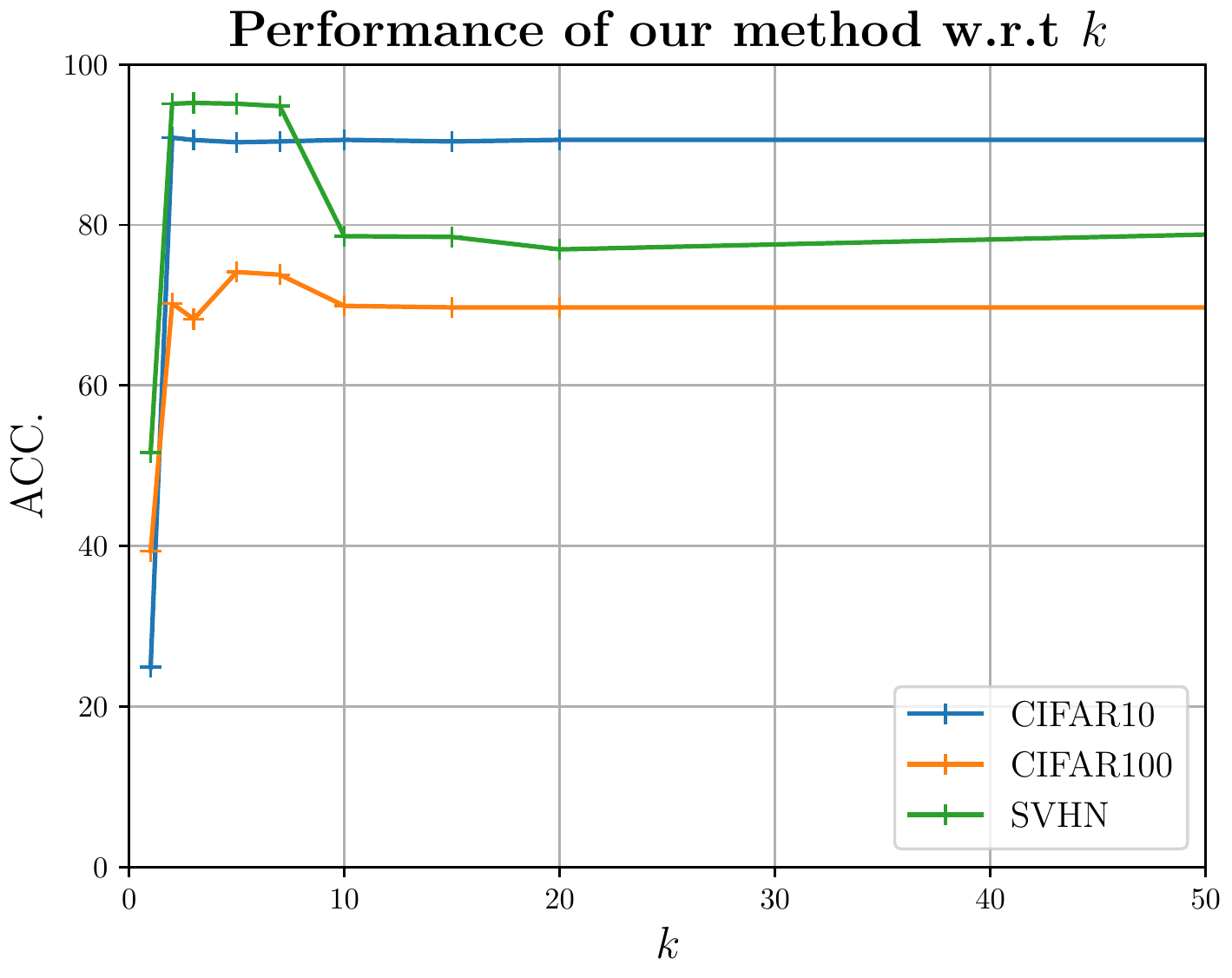}
    \caption{Performance evolution w.r.t. $k$ for ranking statistics. We report results for $k=\{1,2,3,5,7,10,15,20,50\}$.}
    \label{fig:k}
\end{figure}

We also evaluate the evolution of performances of our method with respect to $k$ for ranking statistics. The results on SVHN/CIFAR10/CIFAR100 are shown in \cref{fig:k}. We found that $k=\{5,7\}$ give the best results overall. We also found that for all values of $k$ except 1 results are in general stable.

\subsection{Novel category discovery}

\begin{table}[tb]
 \footnotesize
  \caption{Novel category discovery results on CIFAR10, CIFAR100, and SVHN. ACC on the unlabelled set.
``S.S.'' means self-supervision; \rev{``I.L.'' means incremental learning.}}\label{tab:comparison}
  \centering
  \resizebox{0.48\textwidth}{!}{
  \begin{tabular}{clccc}
    \toprule
    No   &                                      & CIFAR10        & CIFAR100       & SVHN \\
    \midrule
    (1)  & $k$-means~\cite{MackQueen67_Kmeans} & 65.5$\pm$0.0 \%          & 56.6$\pm$1.6\%          & 42.6\%$\pm$0.0\\
    (2)  & KCL~\cite{Hsu18_L2C}                & 66.5$\pm$3.9\%          & 14.3$\pm$1.3\%          & 21.4\%$\pm$0.6 \\
    (3)  & MCL~\cite{Hsu19_MCL}                & 64.2$\pm$0.1\%          & 21.3$\pm$3.4\%          & 38.6\%$\pm$10.8 \\
    (4)  & DTC~\cite{han2019learning}          & 87.5$\pm$0.3\%          & 56.7$\pm$1.2\%          & 60.9\%$\pm$1.6\\
    \midrule
    (5)  & $k$-means~\cite{MackQueen67_Kmeans} w/ S.S.
         & 72.5$\pm$0.0\%                               & 56.3$\pm$1.7\%          & 46.7$\pm$0.0\% \\
    (6)  & KCL~\cite{Hsu18_L2C} w/ S.S.
         & 72.3$\pm$0.2\%                               & 42.1$\pm$1.8\%          & 65.6$\pm$4.9\%\\
    (7)  & MCL~\cite{Hsu19_MCL} w/ S.S.
         & 70.9$\pm$0.1\%                               & 21.5$\pm$2.3\%          & 53.1$\pm$0.3\% \\
    (8)  & DTC~\cite{han2019learning} w/ S.S.
         &88.7$\pm$0.3\%  & 67.3$\pm$1.2\%      & 75.7$\pm$0.4\% \\
\midrule
    (9)  & Ours                                 & \textbf{90.4$\pm$0.5}\% & \textbf{73.2$\pm$2.1}\% & \textbf{95.0$\pm$0.2}\% \\
    (10) & Ours w/ I.L.
         & \textbf{91.7$\pm$0.9}\%                      & \textbf{75.2$\pm$4.2}\% & \textbf{95.2$\pm$0.2}\% \\
    \bottomrule
  \end{tabular}
}
\end{table}

We compare AutoNovel to baselines and state-of-the-art methods for new class discovery, starting from CIFAR10, CIFAR100, and SVHN in~\cref{tab:comparison}.
The first baseline (row 5 in~\cref{tab:comparison}) amounts to applying $k$-means~\cite{MackQueen67_Kmeans} to the features extracted by the fine-tuned model (the second step in~\cref{s:slefsup}), for which we use the $k$-means++ \cite{Arthur2008kmeanspp} initialization.
The second baseline (row 1 in~\cref{tab:comparison}) is similar, but uses as feature extractor a model  trained from scratch using only the labelled images, which corresponds to a standard transfer learning setting.
By comparing rows 1, 5 and 9 in~\cref{tab:comparison}, we can see that our method substantially outperforms $k$-means.
Next, we compare with the KCL~\cite{Hsu18_L2C}, MCL~\cite{Hsu19_MCL} and DTC~\cite{han2019learning} methods.
By comparing rows 2--4 to 9, we see that our method outperforms these by a large margin.
We also try to improve KCL, MCL and DTC by using the same self-supervised initialization we adopt (\cref{s:slefsup}), which indeed results in an improvement (rows 2--4 vs 6--8).
However, their overall performance still lags behind ours by a large margin.
For example, our method of~\cref{s:consistency} achieves $95.0\%$ ACC on SVHN, while ``KCL w/ S.S.'', ``MCL w/ S.S.'' and ``DTC w/ S.S.'' achieve only $65.6\%$, $53.1\%$ and $75.7\%$ ACC, respectively.
Similar trends hold for CIFAR10 and CIFAR100.
Finally, the incremental learning scheme of~\cref{s:incremental} results in further improvements, as can be seen by comparing rows 9 and 10 of~\cref{tab:comparison}.

\begin{figure}[t]
\centering
\tabcolsep=0.02cm
\renewcommand{\arraystretch}{0.25}
\begin{tabular}[b]{ccc}
{\includegraphics[height=6.5em,clip,trim=4cm 0 6cm 1cm]{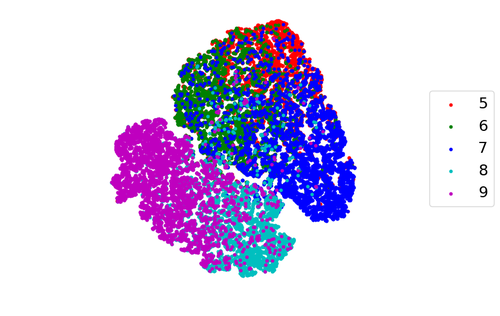}} &
{\includegraphics[height=6.5em,clip,trim=2cm 0 5cm 1cm]{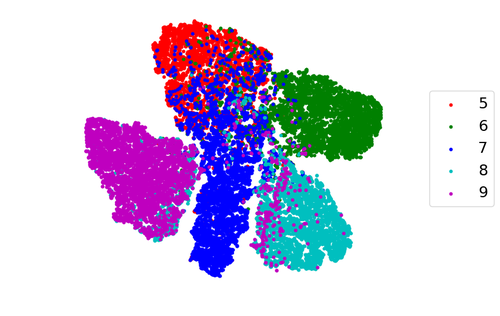}}&
{\includegraphics[height=6.5em,clip,trim=3cm 0 0cm 1cm]{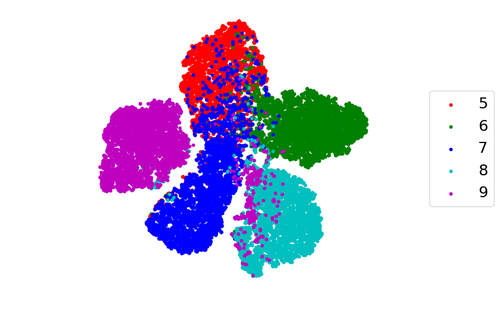}} \\[-0.5em]
(a) init & (b) epoch 30 & (c) epoch 90
\end{tabular}\hfill%
\caption{Evolution of the t-SNE during the training of CIFAR-10. Performed on unlabelled data (i.e., instances of dog, frog, horse, ship, truck).
Colors of data points denote their ground-truth labels.}\label{fig:tsne}
\end{figure}

In~\cref{fig:tsne}, we show the evolution of the learned representation on the unlabelled data from CIFAR10 using t-SNE~\cite{Maaten2008visualizing}.
As can be seen, while the clusters overlap in the beginning, they become more and more separated as the training progresses, showing that our model can effectively discover novel visual categories without labels and learn meaningful embeddings for them.

\begin{table}[t]
 \centering
\caption{Novel category discovery results on OmniGlot and ImageNet. ACC on the unlabelled set.
}\label{tab:comparison_omniglot_imagenet}
  \begin{tabular}[c]{clccc}
    \toprule
    No  &                                      & OmniGlot        & ImageNet \\
    \midrule
    (1) & $k$-means~\cite{MackQueen67_Kmeans} & 77.2\%          & 71.9\%   \\
    (2) & KCL~\cite{Hsu18_L2C}                & 82.4\%          & 73.8\%  \\
    (3) & MCL~\cite{Hsu19_MCL}                & 83.3\%          & 74.4\%  \\
    (4) & DTC~\cite{han2019learning}          & 89.0\%          & 78.3\% \\
    \midrule
    (5) & Ours                                 & \textbf{89.1}\% & \textbf{82.5}\% \\
    \bottomrule
  \end{tabular}
  \hfill

\end{table}

We further compare AutoNovel to others on two more challenging datasets, OmniGlot and ImageNet, in~\cref{tab:comparison_omniglot_imagenet}.
For OmniGlot, results are averaged over the 20 alphabets in the \emph{evaluation} set; for ImageNet, results are averaged over the three 30-class unlabelled sets used in~\cite{Hsu18_L2C,Hsu19_MCL}.
Since we have a relatively larger number of labelled classes in these two datasets, we follow~\cite{han2019learning} and use metric learning on the labelled classes to pre-train the feature extractor, instead of the self-supervised learning.
We empirically found that self-supervision does not provide obvious gains for these two datasets.
This is reasonable since the data in the labelled sets of these two datasets are rather diverse and abundant, so metric learning can provide good feature initialization as there is less class-specific bias due to the large number of pre-training classes.
However, by comparing rows 1 and 5 in~\cref{tab:comparison_omniglot_imagenet}, it is clear that metric learning alone is not sufficient for the task of novel category discovery.
Our method substantially outperforms the $k$-means results obtained using the features from metric learning --- by $11.9\%$ and $10.6\%$ on OmniGlot and ImageNet respectively.
Our method also substantially outperforms the current state-of-the-art, achieving $89.1\%$ and $82.5\%$ ACC on OmniGlot and ImageNet respectively, compared with $89.0\%$ and $78.3\%$ of~\cite{han2019learning}, thus setting the new state-of-the-art.
\rev{By comparing~\cref{tab:comparison} and~\cref{tab:comparison_omniglot_imagenet}, we observe that KCL and MCL perform better on the more challenging ImageNet than the smaller datasets CIAR10, CIFAR100 and SVHN. This can be explained by the fact that the pairwise psuedo labels are provided by a similarity prediction network (SPN) which is pretrained on the labelled data. As there are much less labelled data in CIFAR10, CIFAR100 and SVHN than ImageNet, the learned SPN is less reliable, thus resulting in relatively poor performance for novel category discovery on unlabelled data from new classes.}

\begin{table*}[htb]
\footnotesize
\centering
\caption{Incremental Learning with the novel categories. ``old'' refers to the ACC on the labelled classes while ``new'' refers to the unlabelled classes in the \emph{testing set}.
``all'' indicates the whole testing set.
It should be noted that the predictions are not restricted to their respective subset. ``S.S.'' means self-supervision; \rev{``I.L.'' means incremental learning.}}\label{tab:increment}
\begin{tabular}{lccccccccc}
\toprule
& \multicolumn{3}{c}{CIFAR10} & \multicolumn{3}{c}{CIFAR100} & \multicolumn{3}{c}{SVHN}\\
\cmidrule(rl){2-4}
\cmidrule(rl){5-7}
\cmidrule(rl){8-10}
Classes       & old      & new     & all        & old       & new       & all       & old       & new       & all \\\midrule
KCL w/ S.S.
  & 79.4$\pm$0.6\%   & 60.1$\pm$0.6\%  & 69.8$\pm$0.1\%     & 23.4$\pm$0.3\%    & 29.4$\pm$0.3\%    & 24.6$\pm$0.2\%    & 90.3$\pm$0.3\%    & 65.0$\pm$0.5\%    & 81.0$\pm$0.1\% \\
MCL w/ S.S.
  & 81.4$\pm$0.4\%   & 64.8$\pm$0.4\%  & 73.1$\pm$0.1\%     & 18.2$\pm$0.3\%    & 18.0$\pm$0.1\%    & 18.2$\pm$0.2\%    & 94.0$\pm$0.2\%    & 48.6$\pm$0.3\%    & 77.2$\pm$0.1\% \\
DTC w/ S.S.
  & 58.7$\pm$0.6\%   & 78.6$\pm$0.2\%  & 68.7$\pm$0.3\%     & 47.6$\pm$0.2\%    & 49.1$\pm$0.2\%    & 47.9$\pm$0.2\%    & 90.5$\pm$0.3\%    & 72.8$\pm$0.2\%    & 84.0$\pm$0.1\% \\
\midrule
Ours w/ I.L.
 & \textbf{90.6$\pm$0.2}\% & \textbf{88.8$\pm$0.2}\%  & \textbf{89.7$\pm$0.1}\% & \textbf{71.2$\pm$0.1}\% & \textbf{56.8$\pm$0.3}\% & \textbf{68.3$\pm$0.1}\% & \textbf{96.3$\pm$0.1}\% & \textbf{96.1$\pm$0.0}\% & \textbf{96.2$\pm$0.1}\% \\
\bottomrule
\end{tabular}
\end{table*}

\begin{figure}[t]
\centering
\tabcolsep=0.2em
\renewcommand{\arraystretch}{0.25}
\begin{tabular}[b]{cc}
{\includegraphics[height=7em,clip,trim=0pt 0pt 5cm 0cm]{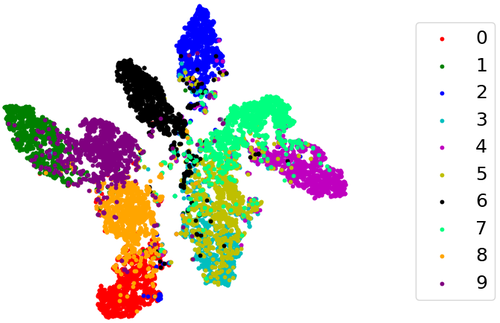}} &
{\includegraphics[height=7em,clip,trim=0pt 0pt 0cm 0cm]{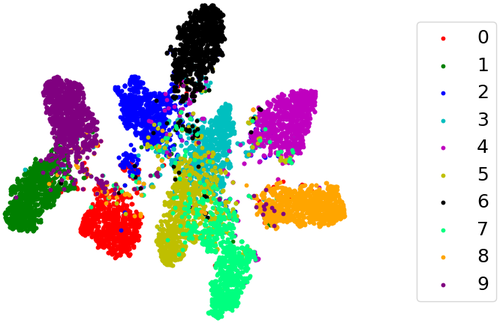}} \\
(a) Ours & (b) +~incr.~learning
\end{tabular}\hfill
\caption{t-SNE on CIFAR10: impact of incremental Learning.
Colors of data points denote their ground-truth labels (``old'' classes 0-4; ``new'' classes 5-9). We observe a bigger overlap in (a) between the ``old'' class 3 and the ``new'' class 5 when not incorporating Incremental Learning.}\label{fig:tsne_both}
\end{figure}

\subsection{Incremental learning scheme}
Here, we further evaluate our incremental scheme for novel category discovery as described in~\cref{s:incremental}.
Methods for novel category discovery such as~\cite{han2019learning, Hsu19_MCL, Hsu18_L2C} focus on obtaining the highest clustering accuracy for the new unlabelled classes, but may forget the existing labelled classes in the process.
In practice, forgetting is not desirable as the model should be able to recognize both old and new classes.
Thus, we argue that the classification accuracy on the labelled classes should be assessed as well, as for any incremental learning setting.
Note however that our setup differs substantially from standard incremental learning~\cite{rebuffi2017icarl,lopez2017gradient,shmelkov2017incremental,aljundi2018memory} where every class is labelled and the focus is on using limited memory.
In our case, we can store and access the original data without memory constraints, but the new classes are unlabelled, which is often encountered in practical applications.

By construction (\cref{s:incremental}), our method learns the new classes on top of the old ones incrementally, out of the box.
In order to compare AutoNovel to methods such as KCL, MCL and DTC that do not have this property, we proceed as follows.
First, the method runs as usual to cluster the unlabelled portion of the data, thus obtaining pseudo-labels for it, and learning a feature extractor as a byproduct.
Then, the feature extractor is used to compute features for both the labelled and unlabelled training data, and a linear classifier is trained using labels and pseudo-labels, jointly on all the classes, old and new.

\begin{figure}
\centering
\tabcolsep=0.02cm
\renewcommand{\arraystretch}{0.25}
\begin{tabular}[b]{cc}
    {\includegraphics[width=0.5\linewidth]{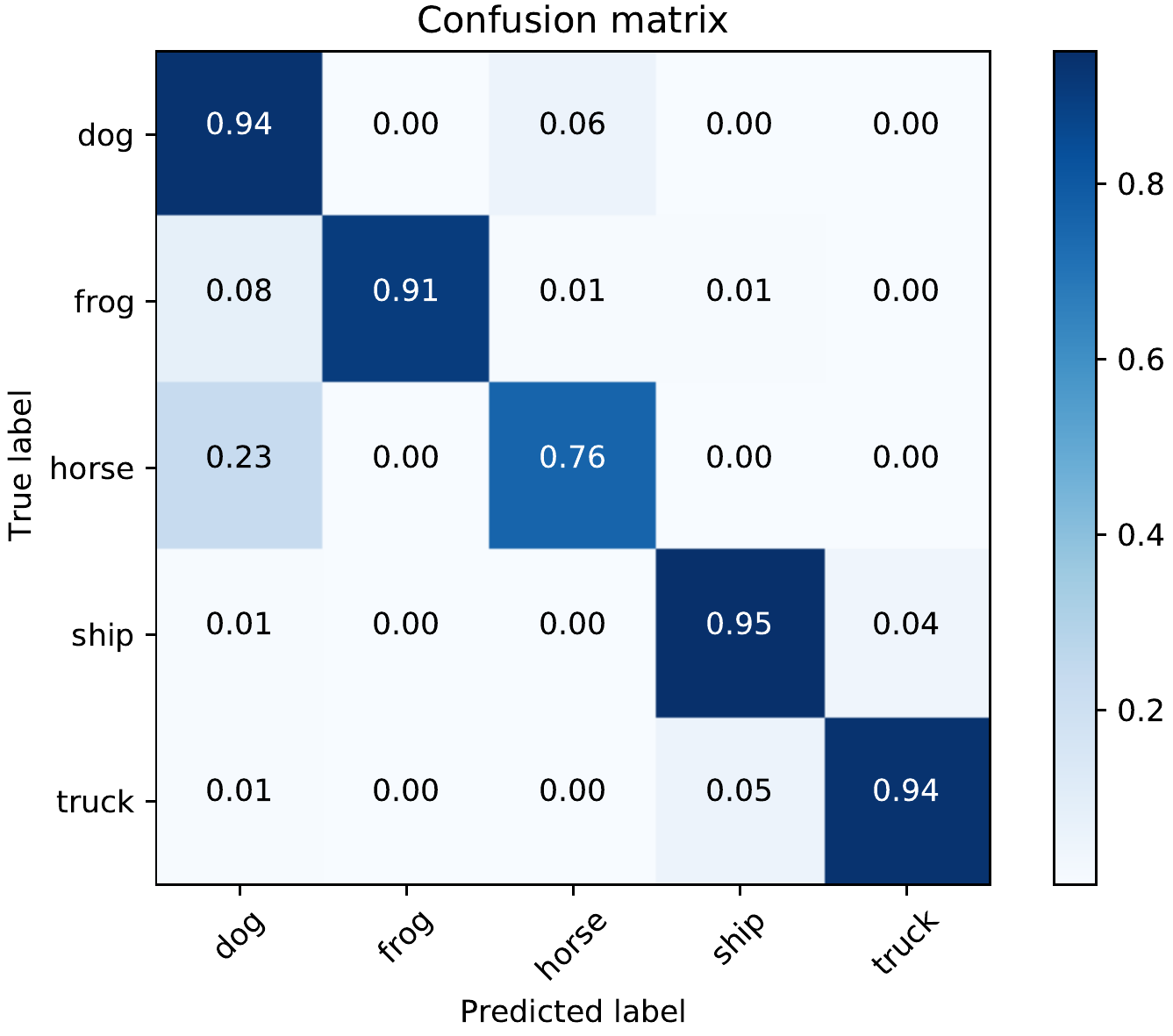}} &
    {\includegraphics[width=0.5\linewidth]{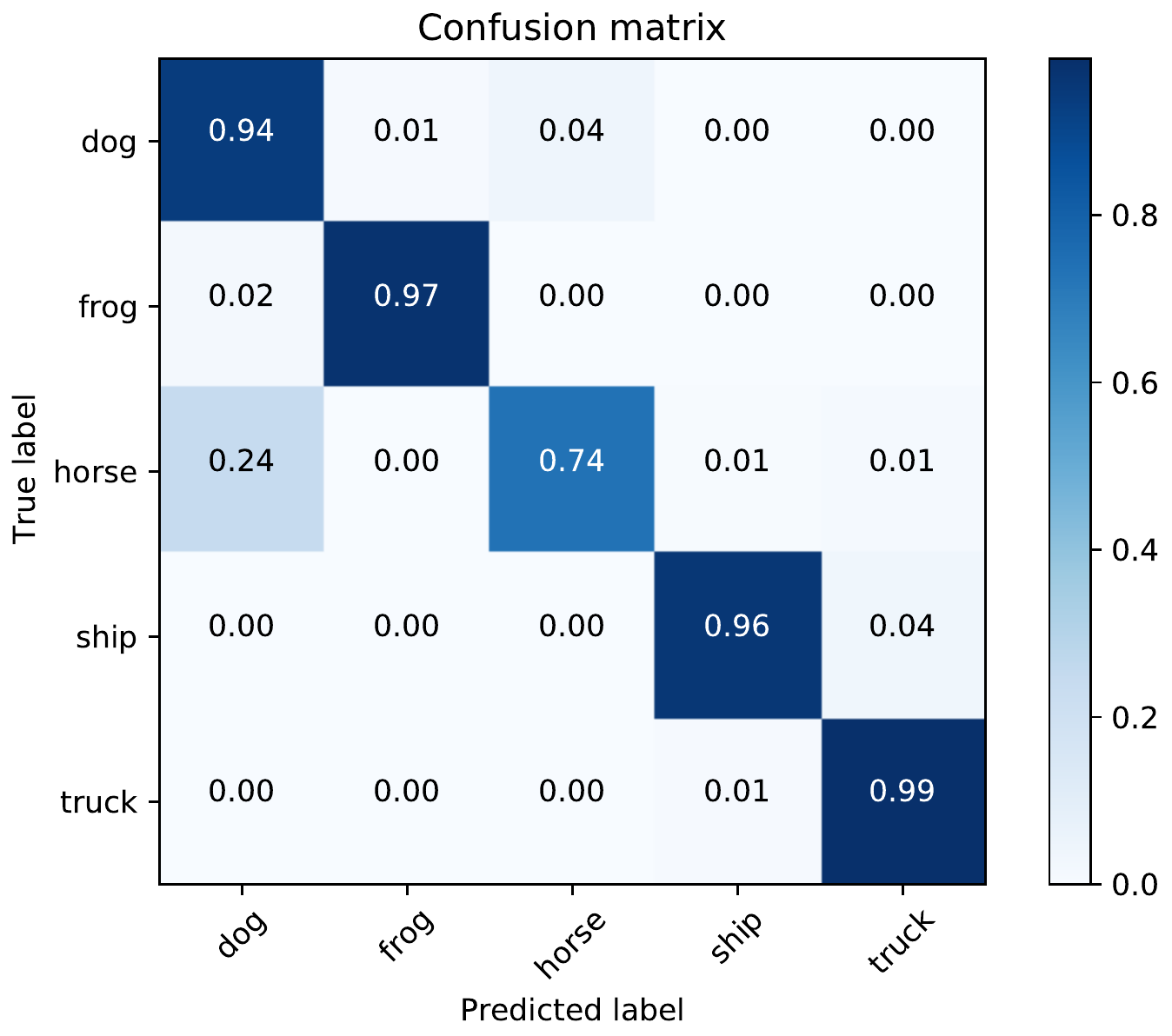}} \\
\end{tabular}\hfill%
\caption{Confusion matrix on unlabelled classes of CIFAR10. Left: our method; Right: our method w/ I.L.}
\label{fig:conf_cifar5}
\end{figure}

We report in~\cref{tab:increment} the performance of the resulting joint classifier networks \emph{on the testing set} of each dataset (this is now entirely disjoint from the training set).
Our method has similar performances on the old and new classes for CIFAR10 and SVHN, as might be expected as the split between old and new classes is balanced.
In comparison, the feature extractor learned by KCL and MCL works much better for the old classes (e.g., the accuracy discrepancy between old and new classes is $25.3\%$ for KCL on SVHN).
Conversely, DTC learns features that work better for the new classes, as shown by the poor performance for the old classes on CIFAR10.
Thus, KCL, MCL and DTC learn representations that are biased to either the old or new classes, resulting overall in suboptimal performance.
In contrast, our method works well on both old and new classes; furthermore, it drastically outperforms existing methods on both.
In~\cref{fig:tsne_both}, we show the t-SNE projection of the learned feature representation on both old and new classes. It can be seen, with incremental learning, the embedding becomes more discriminative between old and new classes. Similarly, in~\cref{fig:conf_cifar5} we compare the confusion matrices w/ and w/o the incremental learning scheme. It can be seen that, with the incremental learning scheme, the clusters for new classes turn out to be more accurate. We notice that the errors are mainly due to the confusion between dog and horse. By looking into the images, we found that images of dogs and horses are confused because of having similar colors or poses.

\subsection{Finding the number of novel categories}
\label{s:results_unknown_k}
\begin{figure}
  \centering
  \includegraphics[width=0.8\linewidth]{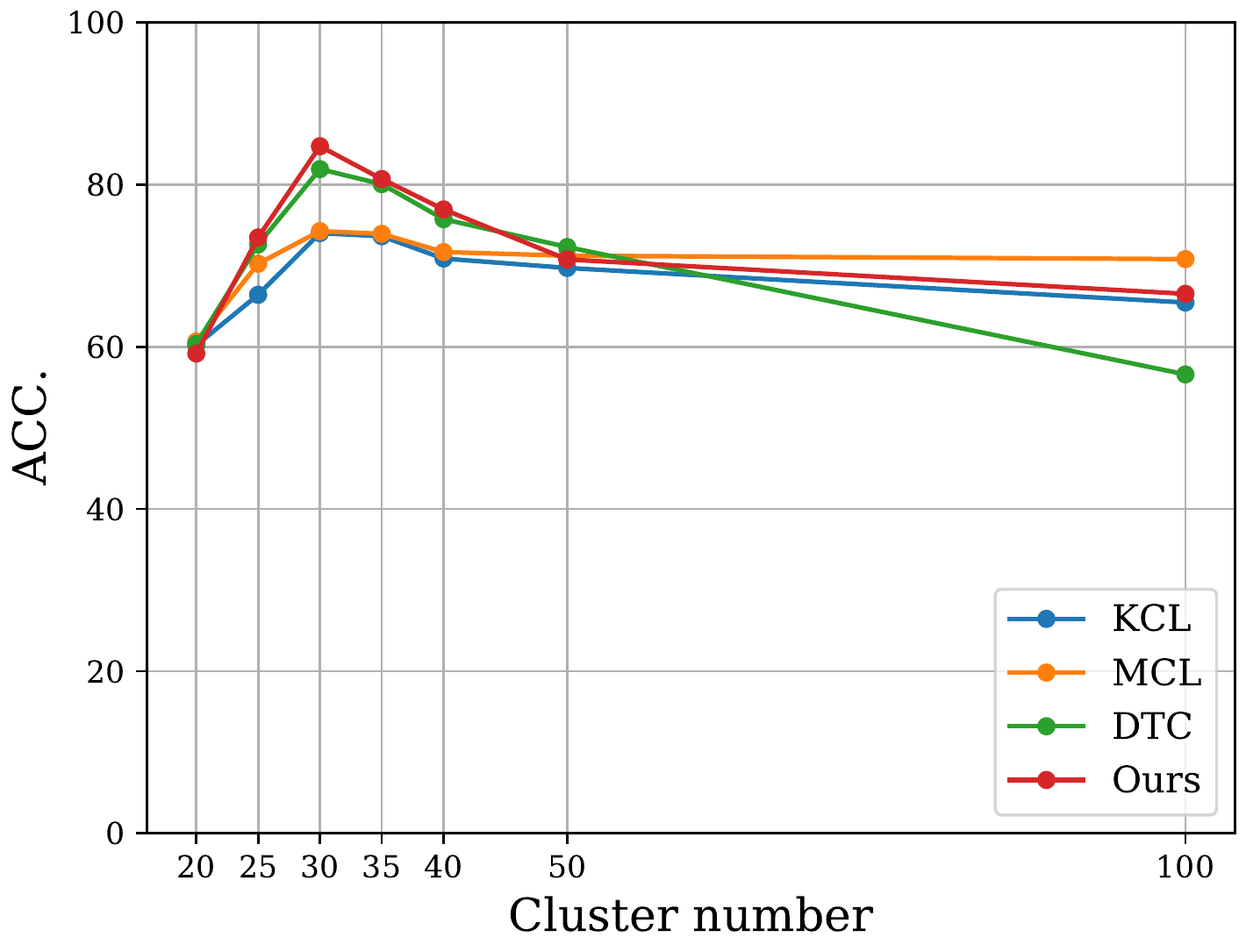}
  \caption{Performance of different methods with different cluster number on ImageNet$_\text{A}$. The ground truth is 30. We vary the cluster number from 20 to 100.}
  \label{fig:cluster_number}
\end{figure}

We now experiment under the more challenging (and realistic) scenario where the number of categories in the unlabelled data is unknown.
KCL and MCL assume the number of categories to be a large value (i.e., 100) instead of estimating the number of categories explicitly.
By contrast, we choose to estimate the number of categories as described in method summary~\ref{alg:unknown_k} (with $C^u_\text{max} = 100$ for all our experiments), before running the transfer clustering algorithm, and only then apply our ranking based method to learn the representation and find the cluster assignment.
Results for novel category number estimation are reported in~\cref{tab:omniglot_k_est} and ~\cref{tab:imagenet_k_est} on OmniGlot and ImageNet respectively. The average errors are 4.6 on OmniGlot and 2.33 on ImageNet, which validates the effectiveness of our approach.
In~\cref{tab:comparison_unknown}, we show the clustering results on OmniGlot and ImageNet, by substituting these estimates to our ranking based method for novel category discovery, and also compare with other methods.
The results of traditional methods are those reported in~\cite{Hsu19_MCL} using raw images for OmniGlot and pretrained features for ImageNet.
AutoNovel outperforms the previous state-of-the-art MCL by 5.2\% and  9.0\% ACC on OmniGlot and ImageNet respectively.
We also experiment on KCL, MCL and DTC by using our estimated number of clusters.
With this augmentation, both KCL amd MCL improve significantly, indicating that our category number estimation method can also be beneficial for other methods. DTC slightly outperforms our ranking based method by $1.6\%$ on OmniGlot with our estimated category number, while our method outperforms DTC by $2.9\%$ on ImageNet.
\rev{In addition, we also validate the sensitivity of different methods to the choice of cluster number on the 30-class ImageNet$_\text{A}$. We vary the cluster number from 20 to 100.  The results are shown in \cref{fig:cluster_number}. It can be seen that the sensitivity to the cluster number is similar for all methods. All methods achieve the best performance when the cluster number equals the ground truth, while the performance drops when the cluster number is off the ground truth. Our method consistently outperforms all others for cluster numbers 25 to 40 (note that our estimated cluster number is 34). For the extreme case with the cluster number of 100, MCL performs the best.}

\begin{table}[ht]
\footnotesize
\centering
\caption{Category number estimation on OmniGlot.}\label{tab:omniglot_k_est}
\setlength\tabcolsep{1pt}
\begin{tabular}{lccccc}
\toprule
Alphabet      & GT & SKMS~\cite{Anand14_SKMS} & KCL~\cite{Hsu18_L2C} & MCL~\cite{Hsu19_MCL}  & Ours \\
\midrule
Angelic & 20 & 16 & 26 & 22 & 23\\
Atemayar Q. & 26 & 17 & 34 & 26 & 25\\
Atlantean & 26 & 21 & 41 & 25 & 34\\
Aurek\_Besh & 26 & 14 & 28 & 22 & 34\\
Avesta & 26 & 8 & 32 & 23 & 31\\
Ge\_ez & 26 & 18 & 32 & 25 & 31\\
Glagolitic & 45 & 18 & 45 & 36 & 46\\
Gurmukhi & 45 & 12 & 43 & 31 & 34\\
Kannada & 41 & 19 & 44 & 30 & 40\\
Keble & 26 & 16 & 28 & 23 & 25\\
Malayalam & 47 & 12 & 47 & 35 & 42\\
Manipuri & 40 & 17 & 41 & 33 & 39\\
Mongolian & 30 & 28 & 36 & 29 & 33\\
Old Church S. & 45 & 23 & 45 & 38 & 51\\
Oriya & 46 & 22 & 49 & 32 & 33\\
Sylheti & 28 & 11 & 50 & 30 & 22\\
Syriac\_Serto & 23 & 19 & 38 & 24 & 26\\
Tengwar & 25 & 12 & 41 & 26 & 28\\
Tibetan & 42 & 15 & 42 & 34 & 43\\
ULOG & 26 & 15 & 40 & 27 & 33\\
\midrule
Avg$_{error}$ & - & 16.3 & 6.35 & 5.10 & \textbf{4.60} \\
\bottomrule
\end{tabular}
\end{table}

\begin{table}[ht]
\centering
\footnotesize
\caption{Category number estimation results.}\label{tab:imagenet_k_est}
\begin{tabular}{lccc}
\toprule
Data & GT  & Ours & Error \\
\midrule
ImageNet$_\text{A}$ & 30 & 34 & 4\\
ImageNet$_\text{B}$ & 30 & 31 & 1\\
ImageNet$_\text{C}$ & 30 & 32 & 2\\
\midrule
Avg$_{error}$ & - & - & 2.33\\
\bottomrule
\end{tabular}
\end{table}

\begin{table}[ht]
\footnotesize
\centering
\caption{Novel category discovery with an unknown class number $C^u$.}\label{tab:comparison_unknown}
\begin{tabular}{lcccc}
 \toprule
    & OmniGlot
    & ImageNet \\
\midrule
Method & ACC  & ACC  \\
\midrule
$k$-means~\cite{MackQueen67_Kmeans} & 18.9\% & 34.5\%\\
LPNMF~\cite{Cai09_LPNMF}  & 16.3\%  & 21.8\%\\
LSC~\cite{Chen11_LSC}  & 18.0\% & 33.5\% \\
\midrule
KCL~\cite{Hsu18_L2C} & 78.1\% & 65.2\% \\
MCL~\cite{Hsu19_MCL} & 80.2\% & 71.5\% \\
\midrule
KCL~\cite{Hsu18_L2C} w/our $C^u$               & 80.3\%          & 71.4\%  \\
MCL~\cite{Hsu19_MCL} w/our $C^u$               & 80.5\%          & 72.9\%  \\
DTC~\cite{han2019learning} w/our $C^u$        & \textbf{87.0}\%          & 77.6\% \\
\midrule
Ours           & 85.4\% & \textbf{80.5}\% \\
\bottomrule
\end{tabular}
\end{table}

\subsection{Transferring from ImageNet pretrained model}
\label{s:results_transfer_imagenet}
Rather than pretraining the model with self-supervised learning, one may also think to transfer representation learned from other datasets. The most common way of transfer learning with modern deep convolutional neural networks is to use ImageNet pretrained models. Here, we explore the potential of leveraging the ImageNet pretrained model to transfer features for novel category discovery. In particular, we take the ImageNet pretrained model as our feature extractor, and finetune the last macro-block and the linear heads of the model using our ranking based method. We experiment on CIFAR10, CIFAR100, and SVHN. The results are shown in \cref{tab:imagenet_transfer}. As can be seen, with the ImageNet pretrained representation, the performance of our method on CIFAR10 and CIFAR100 are further improved w.r.t to the results in~\cref{tab:comparison}. The incremental learning scheme succesfully boosts the performance by $0.7\%$ and $3.8\%$ on CIFAR10 and CIFAR100 respectively. 
\rev{For the performance on SVHN, we notice a significant drop between~\cref{tab:comparison} and~\cref{tab:imagenet_transfer}. }
This is likely due to the small correlation between ImageNet and SVHN, as also noted in other works that try to transfer features from ImageNet to other datasets~\cite{han2019learning,oliver2018realistic}.

\begin{table}[h!]
\centering
\footnotesize
\caption{Transferring from ImageNet to CIFAR10/CIFAR100/SVHN.}\label{tab:imagenet_transfer}
\begin{tabular}{lccc}
\toprule
& CIFAR10 & CIFAR100 & SVHN  \\
\midrule
$k$-means~\cite{MackQueen67_Kmeans} &92.4\%  & 78.8\% & 23.4\% \\
Ours &95.4\% &87.1\% &40.2\% \\
Ours w/I.L. & 96.1\%  & 90.9\% & 38.8\% \\
\bottomrule
\end{tabular}
\end{table}

\subsection{Alternatives to ranking statistics}

\begin{figure}[htb]
    \centering
    \includegraphics[width=0.8\linewidth]{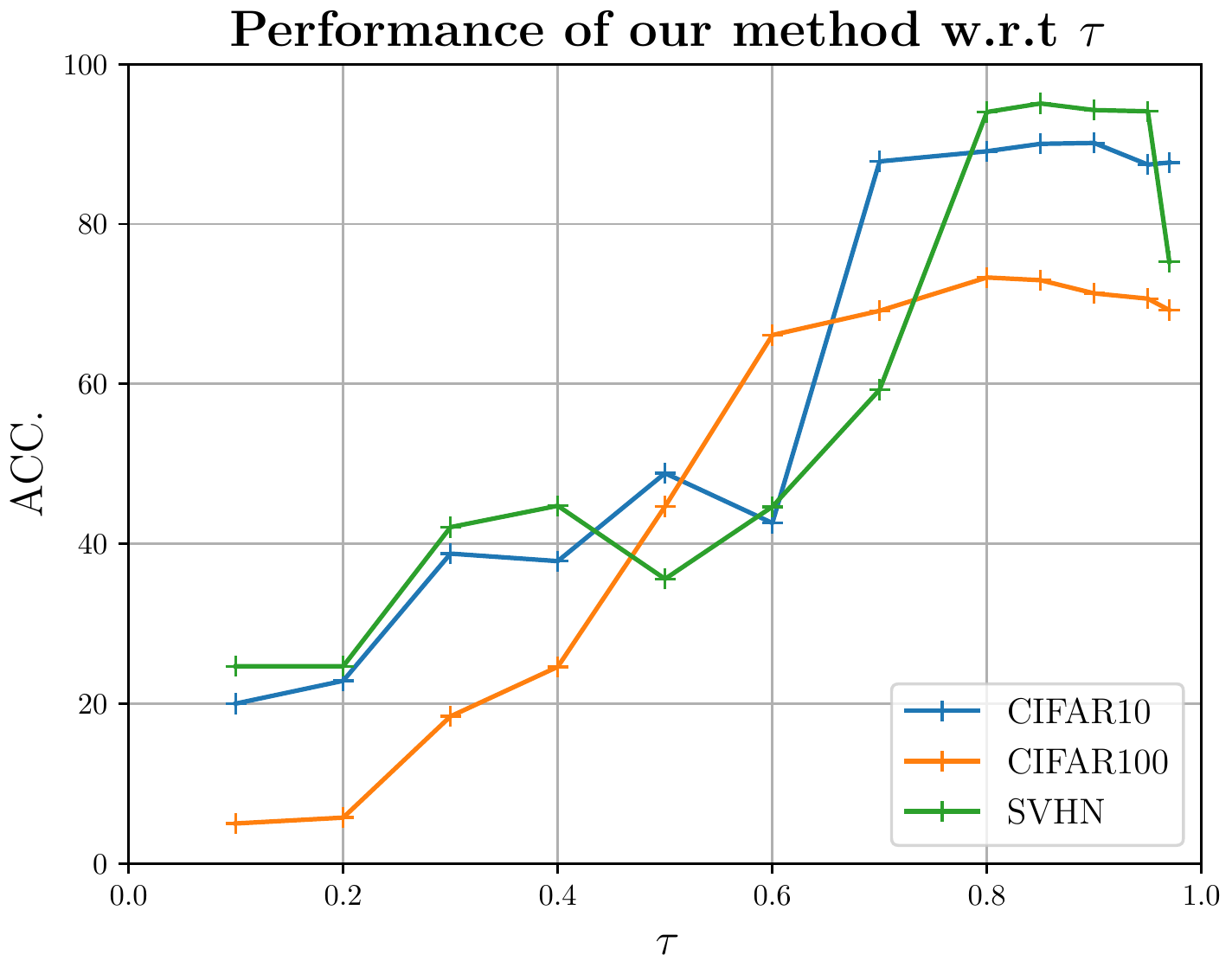}
    \caption{Performance evolution w.r.t. the threshold $\tau$  for cosine similarity. We report results for $\tau=\{0.1,0.2,0.3,0.4,0.5,0.6,0.7,0.8,0.85,0.9,0.95,0.97\}$.}
    \label{fig:cosine}
\end{figure}

The ranking statistics is the key to transfer knowledge from old classes to new classes in our model.
As discussed in~\cref{sec:method:ranking}, other methods like $k$-means, cosine similarity, and nearest neighbor can potentially be used as alternatives to ranking statistics to generate pairwise pseudo labels in AutoNovel.
We experiment with such alternatives and the results are shown in~\cref{tab:diff_pairwise}.
\rev{We experiment on two cases for $k$-means, one by running $k$-means on the mini-batch (denoted as $k$-means (batch)) and the other by running $k$-means (denoted as $k$-means (all)) on the whole unlabelled set.}
As it can be seen, ranking statistics, nearest neighbor and cosine similarity work significantly better than $k$-means on CIFAR10 and SVHN, while ranking statistics and cosine similarity work notably better than nearest neighbor on CIFAR100 and SVHN.
Note that the performance of cosine similarity depends on the choice of a proper threshold $\tau$.
Here, we report the results using the best thresholds on each dataset (0.85/0.8/0.9 for CIFAR10/CIFAR100/SVHN).
The effect of different thresholds is shown in~\cref{fig:cosine}.
It can be seen that, with a carefully chosen threshold, cosine similarity can also be a good measure to generate pairwise pseudo labels in our method, though the results turn to be relatively sensitive to $\tau$.
When $\tau < 0.6$, the cosine similarity fails to provide reliable pairwise pseudo labels.
Meanwhile, as $\tau$ lies in the continues space while $k$ in our ranking based method lies in the discrete integer space, it is easier to set a proper $k$ than $\tau$.
\rev{Overall, while it can be seen that ranking statistics and cosine similarity exhibit a similar behaviour when grid-searching with a relatively low sensitivity to the best value, we still find that ranking statistics is an interesting alternative to cosine similarity and is relatively unexplored in the context of deep learning. Throughout all of our experiments we demonstrate that ranking statistics performs consistently well and could open the way for more applications. Therefore, unless stated otherwise, ranking statistics is our default choice for all experiments.}

\begin{table}[h!]
\centering
\footnotesize
\caption{Different methods for pairwise pseudo labels.}\label{tab:diff_pairwise}
\begin{tabular}{lccc}
\toprule{}
& CIFAR10 & CIFAR100 & SVHN  \\
\midrule
$k$-means (batch) &42.9\%  &\textbf{74.3}\% &45.3\% \\
$k$-means (all) &62.2\%  &55.5\% &61.5\% \\
cosine & 90.1\% &73.3\% & 95.0\% \\
nearest neighbor & 90.2\% &69.7\% & 78.2\% \\
ranking statistics & \textbf{90.4}\%  &73.2\% & 95.0\% \\
\midrule
soft ranking statistics ($k=5$) & 62.2\%  &65.2\% & 72.5\% \\
soft ranking statistics ($k=15$)& 89.7\%  &71.1\% & \textbf{95.2}\% \\
\bottomrule
\end{tabular}
\end{table}

\rev{Through ranking statistics, instead of generating hard (binary) pseudo targets, we can also encode soft rank similarities. To do so, we calculate the shared elements in the top-$k$ rank between two images. Let $c$ be the number of shared elements. The soft similarity is then defined as $\frac{c}{k}$, which can be used to replace the $s_{ij}$ in~\cref{e:bce}. The results are shown in~\cref{tab:diff_pairwise}. We find that $k=5$ is not optimal for the soft rank similarity, as the results largely lag behind the hard counterpart. We adopt the validation method introduced in~\cref{sec:exp:imp} to get $k=15$ as a better choice for the soft rank similarity. The results are in general on par with the hard (binary) rank similarity.}

\subsection{Other self-supervised learning methods}

We adopt the RotNet~\cite{gidaris2018unsupervised} for the first stage of AutoNovel. However, any other self-supervised representation methods can be applied. Here, we further experiment with the latest self-supervised representation learning methods including SimCLR~\cite{chen2020simple}, MoCo~\cite{he2020moco} and MoCo v2~\cite{chen2020mocov2}, which are the state-of-the-art representation learning methods for the tasks of object recognition and detection.
In this experiment, we replace RotNet by the latest self-supervised learning methods in our pipeline, and the other two steps remain the same as before.
In~\cref{tab:diff_selfsup}, we first directly compare the learned feature representations of different methods by running $k$-means on the output of the global average pooling layer on the unlabelled data.
All the three alternatives work better than RotNet when comparing the raw features learned with self-supervised learning on CIFAR10 and CIFAR100, while RotNet performs slightly better on SVHN\@.
This is likely due to nature of the pretext tasks used in these self-supervised learning methods.
RotNet uses the rotation prediction task, which is less relevant to the down stream task of partitioning the unlabelled data based on their semantic meaning.
Differently, the other three methods are contrastive learning based methods, which encourage the images of the same instance to be close in the feature space while the images of different instances to be further away.
Interestingly, we find that RotNet consistently outperforms the other three methods for AutoNovel.
This reveals that better feature initialization does not necessarily mean better representation fine-tuned on downstream tasks like novel category discovery.
Overall, by taking any of these self-supervised learning methods to pre-train our model, the performance can be significantly boosted on novel category discovery by our method.

\begin{table}[htb]
\centering
\footnotesize
\caption{Different self-supervised learning methods.}\label{tab:diff_selfsup}
\begin{tabular}{llccc}
\toprule
& & CIFAR10 & CIFAR100 & SVHN  \\
\midrule
\multirow{4}{*}{$k$-means~\cite{MackQueen67_Kmeans}}
 &SimCLR~\cite{chen2020simple} & \textbf{84.7}\% & \textbf{41.2}\% & 30.6\% \\
 &MoCo~\cite{he2020moco} & 58.7\% & 34.5\% & 21.3\% \\
 &MoCo v2~\cite{chen2020mocov2} & 61.8\% & 39.3\% & 28.5\% \\
 &RotNet~\cite{gidaris2018unsupervised} & 25.5\% &10.3\% & \textbf{31.7}\% \\
\midrule
\multirow{4}{*}{Ours}
 &SimCLR~\cite{chen2020simple} & 89.0\% & 54.6\% & 67.8\% \\
 &MoCo~\cite{he2020moco} & 87.6\% & 61.1\% & 74.6\% \\
 &MoCo v2~\cite{chen2020mocov2} & 89.0\% & 62.5\% & 76.6\% \\
 &RotNet~\cite{gidaris2018unsupervised} & \textbf{90.4}\% & \textbf{73.2}\% & \textbf{95.0}\% \\
\bottomrule
\end{tabular}
\end{table}

\rev{To investigate why RotNet appears to be more effective in our experiments, we carry out experiments by freezing different layers of the network and finetuning the rest of the layers employing different self-supervised learning approaches. ResNet18 is composed of four macro-blocks, denoted as $layer_{\{1, 2, 3, 4\}}$. In~\cref{tab:different_layer}, for each column, we freeze all the parameters before $layer_i$ and finetune $layer_i$ together with the subsequent layers. $head$ denotes the case where we only finetune the two linear heads while $layer_0$ denotes the case where we finetune all the parameters. We measure the novel category performance on CIFAR10 for all methods and report the ACC on the unlabelled data. It can be seen that, if we only finetune the linear heads, SimCLR, MoCo and MoCoV2 significantly outperform RotNet, which is consistent with the conclusion in the literature that the contrastive learning based methods can learn more meaningful higher level feature representation. The higher level features for RotNet is focusing on the task of rotation prediction, which is loosely related to the target task of novel category discovery, thus the performance is poor. However, if we finetune more layers, we can see that the performance are similar, while RotNet appears to be more effective for $layer_4$ and $layer_3$. The strong augmentation is essential for the performance of contrastive learning based self-supervision. However, for SVHN, where multiple digits appear in the same image and only the center digit is to be recognized, strong augmentations like cropping is not suitable for training, because random cropping will change the location of the center digits, which is harmful for training. Therefore, the performance of contrastive learning based methods lags behind RotNet on SVHN as in~\cref{tab:diff_selfsup}.}

\begin{table}[htb]
   \centering
  \caption{Performance on fine-tuning different layers on CIFAR10. ACC on the unlabelled set.
  }\label{tab:different_layer}
  \resizebox{0.48\textwidth}{!}{
    \begin{tabular}[c]{lcccccc}
      \toprule
      Method  & $head$ & $layer_4$ & $layer_3$ & $layer_2$ & $layer_1$ & $layer_0$ \\
      \midrule
      RotNet & 39.9\% & 90.4\% & 90.8\% & 88.4\% & 89.3\% & 88.9\% \\
      SimCLR & 73.1\% & 89.0\% & 89.1\% & 89.5\% & 90.4\% & 88.6\% \\
      MoCo   & 80.8\% & 87.6\% & 88.8\% & 89.3\% & 90.4\% & 89.5\% \\
      MoCoV2 & 84.6\% & 89.0\% & 89.5\% & 89.0\% & 90.3\% & 89.2\% \\
      \bottomrule
    \end{tabular}
}
\end{table}

\subsection{Unsupervised image clustering}
\begin{figure}
\centering
\includegraphics[width=0.75\linewidth]{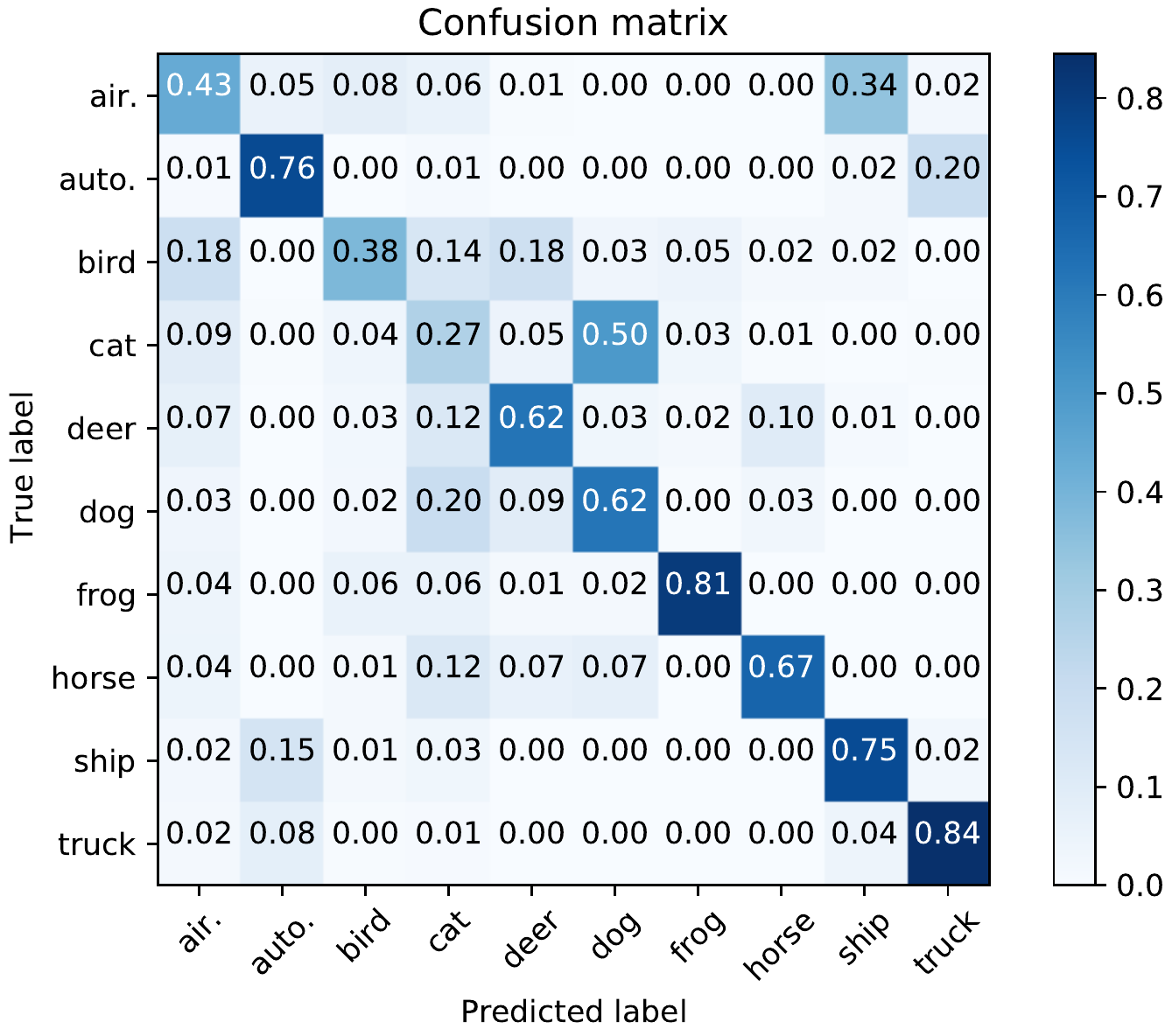}
\caption{Confusion matrix of clustering on CIFAR10.}\label{fig:cm_cluster}
\end{figure}

As discussed in~\cref{s:clustering}, by removing the requirement of labelled data, AutoNovel turns to an unsupervised clustering method that can learn both feature representation and clustering assignment.
Here, we compare our method on the clustering problem with the state-of-the-art methods on three popular benchmarks CIFAR10, CIFAR100 and STL10~\cite{Coates11STL10}.
We follow the common practice to use all 10 classes in CIFAR10 and STL10, and the 20 meta classes in CIFAR100 (denoted as CIFAR100-20) in our experiment for fair comparison.
The results are presented in~\cref{tab:clustering}.
Our method performs on par with the state-of-the-art method IIC~\cite{ji2019invariant} on CIFAR10 and and STL10, while significantly outperforms IIC on CIFAR100-20 by $9.3\%$.
Compared with IIC, which requires extra Sobel filtering on the input data and large batch sizes (660/1000/700 on CIFAR10/CIFAR100-20/STL10), our method only needs the conventional data augmentation (random cropping and horizontal flipping) and a small batch size of 128 for all three datasets.
Therefore, our method is a good alternative to state-of-the-art methods for the task of unsupervised image clustering, though this is not the main objective of this work.
\rev{Moreover, we report the $k$-means results on the feature representation of the base self-supervised model (i.e., RotNet) on each dataset. Unsurprisingly, the results are not satisfactory, because the high level features of RotNet are learned for the task of rotation prediction, making it less effective in capturing useful semantic information for downstream tasks like clustering.}
Meanwhile, we also validate the effectiveness of self-supervised pretraining and consistency regularization in~\cref{eq:cluster_loss}.
We can see that by dropping each of them in our method, the performance drops.
Without the self-supervised pretraining, the performance drops significantly. This suggests that self-supervised learning captures discriminative low level features for the task of image clustering. \rev{Similar to the task of novel category discovery, when applying our approach to unsupervised clustering, the MSE consistency loss is also effective in preventing the ``moving target'' phenomenon described in \cref{s:consistency} during training.}
We show the confusion matrix on CIFAR10 by our full method in~\cref{fig:cm_cluster}.
As can be seen from the diagonal of the matrix, our method can properly cluster objects into proper clusters.
We found airplane and bird are confused with ship because of the shared blue background; cat and dog are confused because of similar poses and colors.

\begin{table}[htb]
\centering
\footnotesize
\caption{Unsupervised image clustering. ``$k$-means on S.S.'' refers to the $k$-means results on the representation of the self-supervised model.}\label{tab:clustering}
\begin{tabular}{lccc}
\toprule{}
& CIFAR10 & CIFAR100-20 & STL10  \\
\midrule
$k$-means~\cite{MackQueen67_Kmeans} &22.9\%  &13.0\% &19.2\% \\
JULE~\cite{yang2016joint} & 27.2\% &13.7\% & 27.7\% \\
DEC~\cite{Xie16_DEC} & 30.1\% &18.5\% & 35.9\% \\
DAC~\cite{Chang_2018_DAC} & 52.2\%  &23.8\% & 47.0\% \\
IIC~\cite{ji2019invariant} &  \textbf{61.7}\%  &25.7\% & \textbf{59.6}\% \\
\midrule
$k$-means on S.S. & 14.3\% & 8.8\% & 15.7\% \\
\midrule
Ours w/o S.S. & 18.8\%  & 13.0\% &  22.7\% \\
Ours w/o MSE &  57.7\%  & 31.6\% &  48.6\% \\
Ours & \textbf{61.7}\%  &\textbf{35.0}\% & 56.4\% \\
\bottomrule
\end{tabular}
\end{table}

\section{Conclusions}\label{sec:conclusion}

In this paper, we have looked at the problem of discovering new classes in an image collection, leveraging labels available for other, known classes.
We have proposed AutoNovel to successfully address this task by combining a few new ideas.
First, the use of self-supervised learning for bootstrapping the image representation trades off the representation quality with its generality, and for our problem this leads to a better solution overall.
Second, we have shown that ranking statistics are an effective method to compare noisy image descriptors, resulting in robust data clustering.
Third, we have shown that jointly optimizing both labelled recognition and unlabelled clustering in an incremental learning setup can reinforce the two tasks while avoiding forgetting.
On standard benchmarks, the combination of these ideas results in much better performance than existing methods that solve the same task.
For larger datasets with more classes and diverse data (e.g., ImageNet) we note that self-supervision can be bypassed as the pretraining on labelled data already provides a powerful enough representation.
In such cases, we still show that the ranking statistics for clustering gives drastic improvement over existing methods.
Besides, we have also proposed a method to estimate the number of categories in the unlabelled data, by transferring knowledge from the labelled data to the unlabelled data, allowing our method to handle the more challenging case when number of categories in unknown.
Finally, we have shown that AutoNovel can also serve as a simple and effective method for unsupervised image clustering by simply removing the requirement of labelled data, performing on par with the state-of-the-art methods.

\rev{One key assumption in our work is that the classes in labelled and unlabelled data follow a similar category definition, in the sense that all classes belong to the same vision dataset. We assume this is the case for categories collected in the same dataset, because we normally follow a consistent procedure to define classes during data curation. 
Ideally, it would be great to have a precise measure about the relevance between labelled and unlabelled tasks so that we can have a clear sense on whether certain algorithms are applicable or not for novel category discovery. We consider this as a potential future research direction.}

\ifCLASSOPTIONcompsoc
  \section*{Acknowledgments}
\else
  \section*{Acknowledgment}
\fi

This  work  is  supported  by  the  EPSRC  Programme  Grant Seebibyte EP/M013774/1, Mathworks/DTA DFR02620, and ERC IDIU-638009. We also gratefully acknowledge the support of Nielsen.

\ifCLASSOPTIONcaptionsoff
  \newpage
\fi

\bibliographystyle{IEEEtran}
\bibliography{IEEEabrv,novel}

\begin{thebibliography}{10}
\providecommand{\url}[1]{#1}
\csname url@samestyle\endcsname
\providecommand{\newblock}{\relax}
\providecommand{\bibinfo}[2]{#2}
\providecommand{\BIBentrySTDinterwordspacing}{\spaceskip=0pt\relax}
\providecommand{\BIBentryALTinterwordstretchfactor}{4}
\providecommand{\BIBentryALTinterwordspacing}{\spaceskip=\fontdimen2\font plus
\BIBentryALTinterwordstretchfactor\fontdimen3\font minus
  \fontdimen4\font\relax}
\providecommand{\BIBforeignlanguage}[2]{{%
\expandafter\ifx\csname l@#1\endcsname\relax
\typeout{** WARNING: IEEEtran.bst: No hyphenation pattern has been}%
\typeout{** loaded for the language `#1'. Using the pattern for}%
\typeout{** the default language instead.}%
\else
\language=\csname l@#1\endcsname
\fi
#2}}
\providecommand{\BIBdecl}{\relax}
\BIBdecl

\bibitem{deng09imagnet}
J.~Deng, W.~Dong, R.~Socher, L.-J. Li, K.~Li, and L.~Fei-Fei, ``Imagenet: A
  large-scale hierarchical image database,'' in \emph{CVPR}, 2009.

\bibitem{oliver2018realistic}
A.~Oliver, A.~Odena, C.~Raffel, E.~D. Cubuk, and I.~J. Goodfellow, ``Realistic
  evaluation of deep semi-supervised learning algorithms,'' in \emph{NeurIPS},
  2018.

\bibitem{Hsu18_L2C}
Y.-C. Hsu, Z.~Lv, and Z.~Kira, ``Learning to cluster in order to transfer
  across domains and tasks,'' in \emph{ICLR}, 2018.

\bibitem{Hsu19_MCL}
Y.-C. Hsu, Z.~Lv, J.~Schlosser, P.~Odom, and Z.~Kira, ``Multi-class
  classification without multi-class labels,'' in \emph{ICLR}, 2019.

\bibitem{han2019learning}
K.~Han, A.~Vedaldi, and A.~Zisserman, ``Learning to discover novel visual
  categories via deep transfer clustering,'' in \emph{ICCV}, 2019.

\bibitem{mccloskey1989catastrophic}
M.~McCloskey and N.~J.Cohen, ``Catastrophic interference in connectionist
  networks: The sequential learning problem,'' \emph{Psychology of Learning and
  Motivation}, 1989.

\bibitem{han20automatically}
K.~Han, S.-A. Rebuffi, S.~Ehrhardt, A.~Vedaldi, and A.~Zisserman,
  ``Automatically discovering and learning new visual categories with ranking
  statistics,'' in \emph{ICLR}, 2020.

\bibitem{chapelle2006semi}
O.~Chapelle, B.~Scholkopf, and A.~Zien, \emph{Semi-Supervised Learning}.\hskip
  1em plus 0.5em minus 0.4em\relax MIT Press, 2006.

\bibitem{rasmus2015semi}
A.~Rasmus, H.~Valpola, M.~Honkala, M.~Berglund, and T.~Raiko, ``Semi-supervised
  learning with ladder networks,'' in \emph{NeurIPS}, 2015.

\bibitem{laine2016temporal}
S.~Laine and T.~Aila, ``Temporal ensembling for semi-supervised learning,'' in
  \emph{ICLR}, 2017.

\bibitem{tarvainen2017mean}
A.~Tarvainen and H.~Valpola, ``Mean teachers are better role models:
  Weight-averaged consistency targets improve semi-supervised deep learning
  results,'' in \emph{NeurIPS}, 2017.

\bibitem{rebuffi2019semi}
S.-A. Rebuffi, S.~Ehrhardt, K.~Han, A.~Vedaldi, and A.~Zisserman,
  ``Semi-supervised learning with scarce annotations,'' \emph{arxiv}, 2019.

\bibitem{48416}
X.~Zhai, A.~Oliver, A.~Kolesnikov, and L.~Beyer, ``S4l: Self-supervised
  semi-supervised learning,'' in \emph{ICCV}, 2019.

\bibitem{sohn2020fixmatch}
K.~Sohn, D.~Berthelot, C.-L. Li, Z.~Zhang, N.~Carlini, E.~D. Cubuk, A.~Kurakin,
  H.~Zhang, and C.~Raffel, ``Fixmatch: Simplifying semi-supervised learning
  with consistency and confidence,'' in \emph{NeurIPS}, 2020.

\bibitem{Pan10transfer}
S.~J. Pan and Q.~Yang, ``A survey on transfer learning,'' \emph{IEEE
  Transactions on Knowledge and Data Engineering}, 2010.

\bibitem{weiss2016asurvey}
K.~Weiss, T.~M. Khoshgoftaar, and D.~Wang, ``A survey of transfer learning,''
  \emph{Journal of Big Data}, 2016.

\bibitem{tan2018asurvey}
C.~Tan, F.~Sun, T.~Kong, W.~Zhang, C.~Yang, and C.~Liu, ``A survey on deep
  transfer learning,'' in \emph{International Conference on Artificial Neural
  Networks}, 2018.

\bibitem{Aggarwal13cluster}
C.~C. Aggarwal and C.~K. Reddy, \emph{Data Clustering: Algorithms and
  Applications}.\hskip 1em plus 0.5em minus 0.4em\relax CRC Press, 2013.

\bibitem{MackQueen67_Kmeans}
J.~MacQueen, ``Some methods for classification and analysis of multivariate
  observations,'' in \emph{Proceedings of the Fifth Berkeley Symposium on
  Mathematical Statistics and Probability}, 1967.

\bibitem{Comaniciu02meanshift}
D.~Comaniciu and P.~Meer, ``Mean shift: A robust approach toward feature space
  analysis.'' \emph{IEEE TPAMI}, 1979.

\bibitem{ng2001onspectral}
A.~Y. Ng, M.~I. Jordan, and Y.~Weiss, ``On spectral clustering: Analysis and an
  algorithm,'' in \emph{NeurIPS}, 2001.

\bibitem{Xie16_DEC}
J.~Xie, R.~Girshick, and A.~Farhadi, ``Unsupervised deep embedding for
  clustering analysis,'' in \emph{ICML}, 2016.

\bibitem{Chang_2017_ICCV}
J.~Chang, L.~Wang, G.~Meng, S.~Xiang, and C.~Pan, ``Deep adaptive image
  clustering,'' in \emph{ICCV}, 2017.

\bibitem{Dizaji2017deepclustering}
K.~G. Dizaji, A.~Herandi, C.~Deng, W.~Cai, and H.~Huang, ``Deep clustering via
  joint convolutional autoencoder embedding and relative entropy
  minimization,'' in \emph{ICCV}, 2017.

\bibitem{Yang17towards}
B.~Yang, X.~Fu, N.~D. Sidiropoulos, and M.~Hong, ``Towards k-means-friendly
  spaces: Simultaneous deep learning and clustering,'' in \emph{ICML}, 2017.

\bibitem{yang2016joint}
J.~Yang, D.~Parikh, and D.~Batra, ``Joint unsupervised learning of deep
  representations and image clusters,'' in \emph{CVPR}, 2016.

\bibitem{hsu2016deep}
Y.-C. Hsu, Z.~Lv, and Z.~Kira, ``Deep image category discovery using a
  transferred similarity function,'' \emph{arxiv}, 2016.

\bibitem{Xian2018zsl}
Y.~Xian, C.~H. Lampert, B.~Schiele, and Z.~Akata, ``Zero-shot learning - a
  comprehensive evaluation of the good, the bad and the ugly,'' \emph{IEEE
  TPAMI}, 2018.

\bibitem{fu2018recent}
Y.~Fu, T.~Xiang, Y.-G. Jiang, X.~Xue, L.~Sigal, and S.~Gong, ``Recent advances
  in zero-shot recognition: Toward data-efficient understanding of visual
  content,'' \emph{IEEE Signal Processing Magazine}, 2018.

\bibitem{Dean2013fast}
T.~Dean, M.~A. Ruzon, M.~Segal, J.~Shlens, S.~Vijayanarasimhan, and J.~Yagnik,
  ``Fast, accurate detection of 100,000 object classes on a single machine,''
  in \emph{CVPR}, 2013.

\bibitem{Yagnik2011thepower}
J.~Yagnik, D.~Strelow, D.~A. Ross, and R.~sung Lin, ``The power of comparative
  reasoning,'' in \emph{ICCV}, 2011.

\bibitem{kolesnikov2019revisiting}
A.~Kolesnikov, X.~Zhai, and L.~Beyer, ``Revisiting self-supervised visual
  representation learning,'' in \emph{CVPR}, 2019.

\bibitem{gidaris2018unsupervised}
S.~Gidaris, P.~Singh, and N.~Komodakis, ``Unsupervised representation learning
  by predicting image rotations,'' \emph{ICLR}, 2018.

\bibitem{chen2020simple}
T.~Chen, S.~Kornblith, M.~Norouzi, and G.~Hinton, ``A simple framework for
  contrastive learning of visual representations,'' in \emph{ICML}, 2020.

\bibitem{he2020moco}
K.~He, H.~Fan, Y.~Wu, S.~Xie, and R.~Girshick, ``Momentum contrast for
  unsupervised visual representation learning,'' in \emph{CVPR}, 2020.

\bibitem{Chang_2018_DAC}
J.~Chang, G.~Meng, L.~Wang, S.~Xiang, and C.~Pan, ``Deep self-evolution
  clustering,'' in \emph{IEEE TPAMI}, 2018.

\bibitem{rebuffi2020lsdc}
S.-A. Rebuffi, S.~Ehrhardt, K.~Han, A.~Vedaldi, and A.~Zisserman, ``Lsd-c:
  Linearly separable deep clusters,'' \emph{arXiv preprint arXiv:2006.10039},
  2020.

\bibitem{Sarfraz19finch}
M.~S. Sarfraz, V.~Sharma, and R.~Stiefelhagen, ``Efficient parameter-free
  clustering using first neighbor relations,'' in \emph{CVPR}, 2019.

\bibitem{kab11onthedistance}
A.~Kab\'{a}n, ``On the distance concentration awareness of certain data
  reduction techniques,'' \emph{Pattern Recognition}, 2011.

\bibitem{rebuffi2017icarl}
S.-A. Rebuffi, A.~Kolesnikov, G.~Sperl, and C.~H. Lampert, ``icarl: Incremental
  classifier and representation learning,'' in \emph{CVPR}, 2017.

\bibitem{lopez2017gradient}
D.~Lopez-Paz and M.~Ranzato, ``Gradient episodic memory for continual
  learning,'' in \emph{NeurIPS}, 2017.

\bibitem{shmelkov2017incremental}
K.~Shmelkov, C.~Schmid, and K.~Alahari, ``Incremental learning of object
  detectors without catastrophic forgetting,'' in \emph{ICCV}, 2017.

\bibitem{aljundi2018memory}
R.~Aljundi, F.~Babiloni, M.~Elhoseiny, M.~Rohrbach, and T.~Tuytelaars, ``Memory
  aware synapses: Learning what (not) to forget,'' in \emph{ECCV}, 2018.

\bibitem{ji2019invariant}
X.~Ji, J.~F. Henriques, and A.~Vedaldi, ``Invariant information clustering for
  unsupervised image classification and segmentation,'' in \emph{ICCV}, 2019.

\bibitem{kuhn1955hungarian}
H.~W. Kuhn, ``The hungarian method for the assignment problem,'' \emph{Naval
  research logistics quarterly}, 1955.

\bibitem{Arbelaitz12cluster}
O.~Arbelaitz, I.~Gurrutxaga, J.~Muguerza, J.~M. P\'{e}rez, and I.~Perona, ``An
  extensive comparative study of cluster validity indices,'' \emph{Pattern
  Recognition}, 2012.

\bibitem{Rousseeuw87Silhouettes}
P.~J. Rousseeuw, ``Silhouettes: A graphical aid to the interpretation and
  validation of cluster analysis,'' \emph{Journal of Computational and Applied
  Mathematics}, 1987.

\bibitem{Dunn74dunn}
J.~C. Dunn, ``Well-separated clusters and optimal fuzzy partitions,''
  \emph{Journal of Cybernetics}, 1974.

\bibitem{Davies79pami}
D.~L. Davies and D.~W. Bouldin, ``A cluster separation measure,'' \emph{IEEE
  TPAMI}, 1979.

\bibitem{Cali74cluster}
T.~Cali\'{n}ski and J.~Harabasz, ``A dendrite method for cluster analysis,''
  \emph{Communications in Statistics-theory and Methods}, 1974.

\bibitem{Bezdek98cluster}
J.~C. Bezdek and N.~R. Pal, ``Some new indexes of cluster validity,''
  \emph{IEEE Transactions on Systems, Man, and Cybernetics, Part B}, 1998.

\bibitem{Krizhevsky09cifar}
A.~Krizhevsky and G.~Hinton, ``Learning multiple layers of features from tiny
  images,'' \emph{Technical report}, 2009.

\bibitem{Netzer2011svhn}
Y.~Netzer, T.~Wang, A.~Coates, A.~Bissacco, B.~Wu, and A.~Y. Ng, ``Reading
  digits in natural images with unsupervised feature learning,'' in
  \emph{NeurIPS Workshop on Deep Learning and Unsupervised Feature Learning},
  2011.

\bibitem{Lake15omnniglot}
B.~M. Lake, R.~Salakhutdinov, and J.~B. Tenenbaum, ``Human-level concept
  learning through probabilistic program induction,'' \emph{Science}, 2015.

\bibitem{vangansbeke2020scan}
W.~Van~Gansbeke, S.~Vandenhende, S.~Georgoulis, M.~Proesmans, and L.~Van~Gool,
  ``Scan: Learning to classify images without labels,'' in \emph{ECCV}, 2020.

\bibitem{he2016deep}
K.~He, X.~Zhang, S.~Ren, and J.~Sun, ``Deep residual learning for image
  recognition,'' in \emph{CVPR}, 2016.

\bibitem{simonyan15vgg}
K.~Simonyan and A.~Zisserman, ``Very deep convolutional networks for
  large-scale image recognition,'' in \emph{ICLR}, 2015.

\bibitem{sutskever2013importance}
I.~Sutskever, J.~Martens, G.~Dahl, and G.~Hinton, ``On the importance of
  initialization and momentum in deep learning,'' in \emph{ICML}, 2013.

\bibitem{kingma2014adam}
D.~P. Kingma and J.~Ba, ``Adam: A method for stochastic optimization,'' in
  \emph{ICLR}, 2014.

\bibitem{Arthur2008kmeanspp}
D.~Arthur and S.~Vassilvitskii, ``k-means++: the advantages of careful
  seeding,'' in \emph{ACM-SIAM symposium on Discrete algorithms}, 2007.

\bibitem{Maaten2008visualizing}
L.~van~der Maaten and G.~Hinton, ``Visualizing data using t-sne,'' \emph{JMLR},
  2008.

\bibitem{Anand14_SKMS}
S.~Anand, S.~Mittal, O.~Tuzel, and P.~Meer, ``Semi-supervised kernel mean shift
  clustering,'' \emph{IEEE TPAMI}, 2014.

\bibitem{Cai09_LPNMF}
D.~Cai, X.~He, X.~Wang, H.~Bao, and J.~Han., ``Locality preserving nonnegative
  matrix factorization,'' in \emph{IJCAI}, 2009.

\bibitem{Chen11_LSC}
X.~Chen and D.~Cai, ``Large scale spectral clustering with landmark-based
  representation,'' in \emph{AAAI}, 2011.

\bibitem{chen2020mocov2}
X.~Chen, H.~Fan, R.~Girshick, and K.~He, ``Improved baselines with momentum
  contrastive learning,'' \emph{arXiv preprint arXiv:2003.04297}, 2020.

\bibitem{Coates11STL10}
A.~Coates, H.~Lee, and A.~Y. Ng, ``An analysis of single layer networks in
  unsupervised feature learning,'' in \emph{AISTATS}, 2011.

\end{thebibliography}

\begin{IEEEbiography}[{\includegraphics[width=1in,height=1.25in,clip,keepaspectratio]{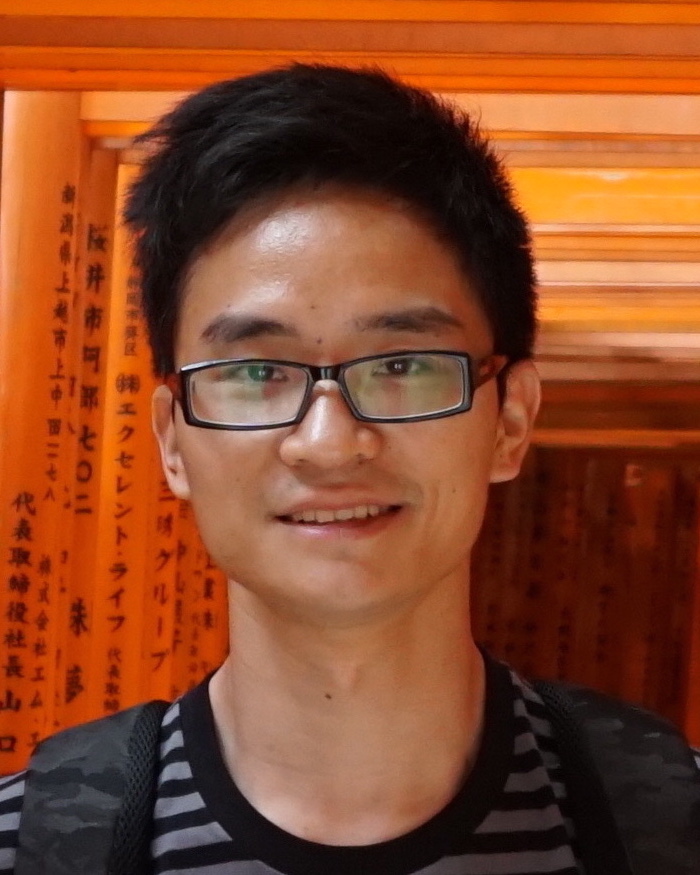}}]{Kai Han} received the Ph.D. degree from the Department of Computer Science, The University of Hong Kong, in 2018. 
He was a Postdoctoral Researcher with the Visual Geometry Group (VGG), Department of Engineering Science, University of Oxford, from 2018 to 2020. 
He is currently a Lecturer (Assistant Professor) in Computer Vision with the Department of Computer Science at University of Bristol. His research interests include computer vision and deep learning.
\end{IEEEbiography}
\vspace{-2em}

\begin{IEEEbiography}[{\includegraphics[width=1in,height=1.25in,clip,keepaspectratio]{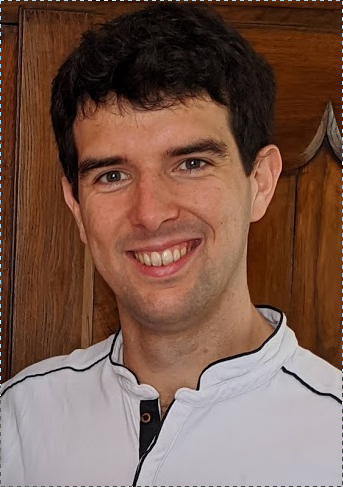}}]{Sylvestre-Alvise Rebuffi} studied at Ecole Centrale Paris, ENS Cachan and then obtained in 2020 his DPhil from the University of Oxford where he was a member of the Visual Geometry Group (VGG) from 2016 to 2020. He is currently a Research Scientist at Google DeepMind. His research spans diverse machine learning and computer vision topics ranging from continual learning, multi-task learning, interpretability and robustness.
\end{IEEEbiography}
\vspace{-2em}

\begin{IEEEbiography}[{\includegraphics[width=1in,height=1.25in,clip,keepaspectratio]{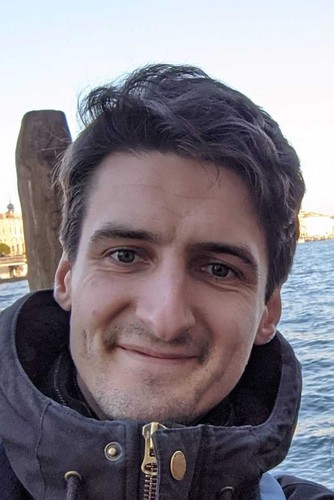}}]{S\'ebastien Ehrhardt} received his DPhil from the University of Oxford in 2021. He is a former member of the Visual Geometry Group (VGG), Department of Engineering Science, University of Oxford, where he worked under Andrea Vedaldi's supervision from 2016 to 2020. 
He is currently an Applied Scientist at Onfido. 
His research interests revolves the application of deep learning to vision datasets with limited number of annotated data. 
\end{IEEEbiography}
\vspace{-2em}

\begin{IEEEbiography}[{\includegraphics[width=1in,height=1.25in,clip,keepaspectratio]{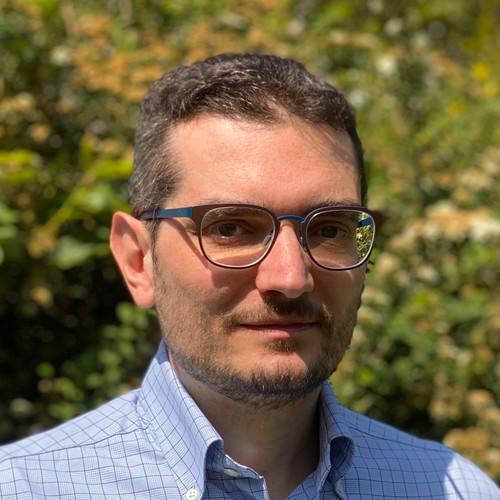}}]{Andrea Vedaldi}
is currently a professor of computer vision and machine learning with the University of Oxford, where he has been co-leading Visual Geometry Group since 2012. He is also a research scientist with Facebook AI Research, London. He has authored or coauthored more than 130 peer-reviewed publications in the top machine vision and artificial intelligence conferences and journals. His research interests include unsupervised learning of representations and geometry in computer vision. He was the recipient of the Mark Everingham Prize for selfless contributions to the computer vision community, the Open Source Software Award by the ACM, and the Best Paper Award from the Conference on Computer Vision and Pattern Recognition.
\end{IEEEbiography}
\vspace{-2em}

\begin{IEEEbiographynophoto}{Andrew Zisserman}
is the Professor of Computer Vision Engineering in the Department of Engineering Science at the University of Oxford.
\end{IEEEbiographynophoto}

\end{document}